\documentclass[journal]{IEEEtran}
\usepackage{amsmath,amsfonts}
\usepackage{algorithmic}
\usepackage{array}
\usepackage[caption=false,font=normalsize,labelfont=sf,textfont=sf]{subfig}
\usepackage{textcomp}
\usepackage{stfloats}
\usepackage{url}
\usepackage{verbatim}
\usepackage{graphicx}
\usepackage{algorithm}
\usepackage{float}
\usepackage[table]{xcolor}
\usepackage{colortbl}
\def\BibTeX{{\rm B\kern-.05em{\sc i\kern-.025em b}\kern-.08em
    T\kern-.1667em\lower.7ex\hbox{E}\kern-.125emX}}
\usepackage{balance}

\usepackage{amssymb}
\usepackage{cite}
\usepackage{booktabs}
\usepackage{multirow}
\usepackage{enumitem}

\usepackage[colorlinks=true,
linkcolor=blue,
citecolor=blue,
urlcolor=blue]{hyperref}
\begin{document}
\title{Reinforcement Learning-Guided Evolutionary Policy Optimization for Preference-Adjustable Heterogeneous Agile Earth Observation Satellite Scheduling}

\author{He Wang, Junyu Wu, Hui Li, Yanjie Song, \IEEEmembership{Senior Member,~IEEE}, \\Witold Pedrycz, \IEEEmembership{Life Fellow,~IEEE}  and Liang Li
	\thanks{This research was jointly funded by the National Natural Science Foundation of China (Nos. 62373117, 62403158, 62573150, 72501042, 62503128), the Fundamental Research Funds for Central Universities (No.3072025GH0401) and the Hainan Provincial Natural Science Foundation of 
		China (No.626QN0688).(Corresponding
		author: Liang Li)

		He Wang, Junyu Wu, Hui Li and Liang Li are with the College of Intelligent Science and Engineering, Harbin Engineering University, Harbin 150001, China (e-mail: wang\_he@hrbeu.edu.cn,  wujunyu@hrbeu.edu.cn, lihuiheu@hrbeu.edu.cn, liliang@hrbeu.edu.cn).

		Yanjie Song is with the School of Information Science and Technology, Dalian Maritime University, Dalian 116026, China(e-mail: songyj\_2017@163.com).

	Witold Pedrycz is with the Department of Electrical and Computer Engineering, University of Alberta, Edmonton, AB T6G 2R3, Canada, the Institute of Systems Engineering, Macau University of Science and Technology, Taipa 999078, Macau SAR, China, the Systems Research Institute, Polish Academy of Sciences, 00-901 Warsaw, Poland, the Research Center of Performance and Productivity Analysis, Istinye University, Istanbul 34396, Türkiye, and Constructor University, Bremen 28759, Germany (e-mail: wpedrycz@ualberta.ca).

	}
\thanks{%
	\textcopyright~2026 IEEE. Personal use of this material is permitted.
	Permission from IEEE must be obtained for all other uses, in any current
	or future media, including reprinting/republishing this material for
	advertising or promotional purposes, creating new collective works,
	for resale or redistribution to servers or lists, or reuse of any
	copyrighted component of this work in other works.
	Digital Object Identifier:
	\href{https://doi.org/10.1109/TGRS.2026.3745218}
	{10.1109/TGRS.2026.3745218}.%
}
}

\markboth{IEEE Transactions on Geoscience and Remote Sensing}%
{Wang \MakeLowercase{\textit{et al.}}: Reinforcement Learning-Guided Evolutionary Policy Optimization}

\maketitle

\begin{abstract}
Heterogeneous agile Earth observation satellite (AEOS) scheduling is a constrained combinatorial optimization problem involving task selection, satellite assignment, observation-window selection, and execution sequencing under satellite-dependent visibility, attitude-transition, energy, storage, and load-distribution constraints.
To address this problem, this paper proposes an Evolutionary Policy Optimization (EPO) framework for preference-adjustable heterogeneous AEOS scheduling.
An assignment-based indirect encoding is combined with a decoder-based equivalent-cost model, which preserves satellite-specific feasibility constraints while integrating task gain, energy saving, and load balance into a weighted utility.
Within this framework, reinforcement learning is used as an online operator-mode controller rather than a direct schedule constructor.
The actor--critic policy selects high-level AEOS-oriented search modes to coordinate exploration, feasibility recovery, diversity maintenance, and local refinement during evolutionary search.
Based on EPO, a reinforcement-learning-assisted operator-selection memetic evolutionary algorithm, termed RLOSMEA, is developed.
Experiments on heterogeneous AEOS scenarios show that RLOSMEA achieves higher weighted utility and more stable convergence than the compared algorithms under the same evaluation budget.
Additional analyses on preference adjustability, parameter robustness, ablation variants, and learning behavior further verify the effectiveness and robustness of the proposed framework.
\end{abstract}

\begin{IEEEkeywords}
Satellite scheduling, Earth observation, reinforcement learning, evolutionary policy optimization, preference-adjustable optimization.
\end{IEEEkeywords}

\section{Introduction}
\label{sec:introduction}

\IEEEPARstart{A}{gile} Earth observation satellites (AEOSs) provide flexible and time-sensitive observation capabilities for applications such as disaster response, environmental monitoring, infrastructure inspection, and wide-area surveillance~\cite{01,02,03,04}.
By performing rapid attitude maneuvers, AEOSs can access geographically distributed targets within flexible visibility windows.
This agility increases observation opportunities, but also makes task scheduling more difficult, especially in heterogeneous multi-satellite systems where orbital access, maneuvering capability, onboard resources, and payload constraints differ across platforms\cite{05}.

\begin{figure*}[!t]
	\centering
	\includegraphics[width=\textwidth]{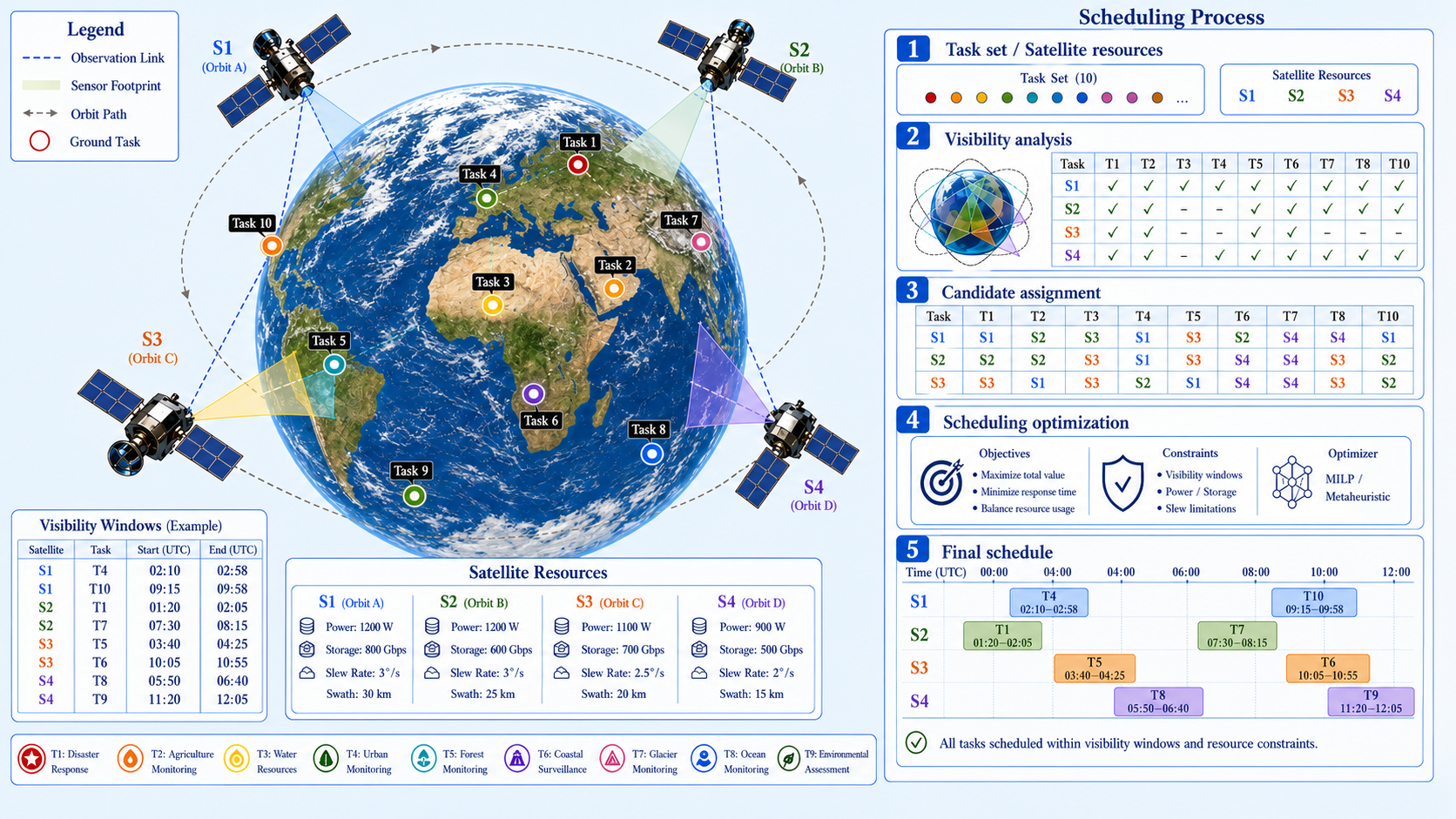}
	\caption{Overall scheduling process for heterogeneous AEOS systems, including visibility analysis, satellite--task assignment, scheduling optimization, and final schedule generation.}
	\label{fig01}
\end{figure*}

Fig.~\ref{fig01} illustrates the overall workflow of heterogeneous AEOS scheduling, where visibility analysis, satellite--task assignment, scheduling optimization, and sequence generation are coupled with temporal, attitude-transition, energy, storage, and load-distribution constraints.

Heterogeneous AEOS scheduling requires the joint determination of task selection, satellite assignment, observation-window selection, and execution sequencing.
For the same task, different satellites may provide different visibility windows, attitude-transition requirements, energy consumption, storage usage, and load contributions.
Therefore, a schedule must satisfy coupled visibility, temporal, attitude-transition, energy, storage, and load-distribution constraints.
These characteristics make heterogeneous AEOS scheduling a strongly constrained combinatorial optimization problem\cite{06}.

Early studies on Earth observation satellite scheduling established the basic modeling and algorithmic foundations of task selection and sequencing under visibility-window and resource constraints~\cite{07,08,09,10}.
Exact optimization methods, including branch-and-bound, dynamic programming, mixed-integer programming, branch-and-price, and branch-and-cut-and-price, can provide high-quality or optimal solutions for medium-scale or specially structured cases~\cite{11,12,13,14}.
However, their computational cost increases rapidly with the numbers of satellites, tasks, candidate windows, and temporal coupling relationships.
This limits their direct use in large-scale or repeatedly updated AEOS scheduling scenarios that require high-quality solutions within a limited computational budget.

Heuristic and metaheuristic algorithms have therefore been widely used in AEOS scheduling and related mission-planning problems.
Greedy insertion, local search, tabu search, adaptive large neighborhood search, genetic algorithms, particle swarm optimization, ant colony optimization, differential evolution, and memetic algorithms can generate feasible and competitive schedules more efficiently than exact methods in large-scale cases~\cite{15,16,17,18,19,20}.
Nevertheless, many existing metaheuristics still rely on fixed operator probabilities, predefined neighborhood structures, or manually designed adaptive rules.
Such mechanisms may become ineffective when the population diversity is low, constraint violations are severe, or rare high-value tasks are insufficiently captured.

Another related direction is multi-objective or preference-oriented satellite scheduling.
Existing studies have optimized observation profit, energy consumption, image quality, failure risk, and load balance using multi-objective formulations and strategy-fusion evolutionary optimization~\cite{22,23,24,25,26}.
These methods are useful for analyzing tradeoffs among different objectives.
However, Pareto-based methods usually return a set of non-dominated schedules, whereas practical mission planning often requires one deployable schedule under a specified operational preference.
Therefore, a preference-adjustable scalar utility model is useful when the individual utility components remain interpretable.

Reinforcement learning (RL) has also been introduced into satellite scheduling to improve adaptive decision making.
Existing studies have modeled AEOS scheduling as a Markov decision process (MDP), explored onboard RL planning, and developed deep reinforcement learning (DRL), attention-based, graph-based, and energy-aware scheduling models~\cite{21,22,27,28,29,30,31,32,33}.
More recently, large language model-agent-enhanced online scheduling has also been investigated for large-scale AEOS systems, where large language model agents are combined with bi-level dispatching and heuristic optimization to support adaptive online decision making~\cite{34}.
These methods demonstrate the potential of learning-based scheduling.
However, direct RL or DRL schedulers usually construct schedules by sequentially selecting tasks, observation windows, or scheduling actions.
They therefore face large discrete action spaces, strict feasibility constraints, sparse or unstable rewards, and high training costs.
Their performance may also depend on the training scenarios and may not generalize reliably when task density, satellite resources, visibility-window structures, or operational preferences change.

Compared with direct RL schedulers, evolutionary and swarm intelligence algorithms provide mature population-based search mechanisms for constrained combinatorial optimization without requiring a pre-trained policy.
However, conventional evolutionary algorithms still require effective control over operator selection, parameter adaptation, feasibility recovery, and local refinement.
RL-assisted evolutionary algorithms have been investigated in broader optimization fields, where RL is commonly used for operator selection, parameter control, subpopulation management, or heuristic coordination~\cite{35,36,37,38}.
These studies suggest that RL can improve the adaptability of evolutionary search without replacing the optimizer itself.
Nevertheless, in AEOS scheduling, the integration of RL and evolutionary optimization remains mainly limited to direct RL schedulers, learning-guided construction methods, or RL-controlled local neighborhood search.
A systematic framework that embeds an online actor--critic policy into memetic evolutionary search for high-level AEOS-specific operator-mode selection remains insufficiently explored.

To address these limitations, this paper develops an Evolutionary Policy Optimization (EPO) framework for preference-adjustable heterogeneous AEOS scheduling.
The proposed framework combines an assignment-based weighted scheduling model, decoder-based equivalent-cost evaluation, population-based evolutionary search, and online actor--critic operator-mode selection.
The policy does not construct schedules directly; instead, it selects high-level AEOS-oriented operator modes that coordinate exploration, feasibility recovery, diversity maintenance, and schedule refinement.
This design keeps the RL action space compact, preserves the feasibility-handling capability of the decoder, and retains the global search capability of evolutionary optimization.

The main contributions of this paper are summarized as follows.

\begin{enumerate}

\item We formulate a preference-adjustable heterogeneous AEOS scheduling model by combining assignment-based indirect encoding, decoder-based schedule construction, and equivalent-cost evaluation.
The model preserves satellite-dependent visibility, attitude-transition, energy, storage, and load characteristics, while integrating task gain, energy saving, and load balance into an interpretable weighted utility.

\item We propose a modular EPO framework that integrates population-based evolutionary search with online actor--critic operator-mode control.
The policy layer selects high-level operator modes instead of directly constructing schedules, thereby keeping the RL action space compact and allowing the decoder to handle feasibility and resource constraints.

\item We develop RLOSMEA, a reinforcement-learning-assisted operator-selection memetic evolutionary algorithm for heterogeneous AEOS scheduling.
The algorithm adaptively coordinates exploration, feasibility recovery, schedule refinement, and diversity maintenance.
Its effectiveness is verified under different heterogeneous scenarios, preference settings, parameter configurations, and ablation variants.
\end{enumerate}

The remainder of this paper is organized as follows.
Section~\ref{sec:model} presents the preference-adjustable heterogeneous AEOS scheduling model.
Section~\ref{sec:framework} describes the EPO framework and the RLOSMEA algorithm.
Section~\ref{sec:experiments} reports the experimental settings, comparative results, preference-adjustability analysis, parameter robustness analysis, ablation study, and learning-behavior analysis.
Section~\ref{sec:conclusion} concludes this paper and discusses future work.

\section{Scheduling Model for Heterogeneous AEOS}
\label{sec:model}

\subsection{Problem Description}

Heterogeneous AEOS scheduling aims to determine task selection, satellite assignment, observation-window selection, and execution sequencing under coupled visibility, temporal, attitude-transition, energy, storage, and load-distribution constraints.
In this study, heterogeneity means that different satellites may provide different access opportunities, payload/resource capabilities, execution costs, and scheduling contributions for the same task.
Orbital heterogeneity is reflected by satellite-dependent visibility windows caused by different orbital altitudes, inclinations, initial phases, and ground-track coverage.
Payload and resource heterogeneity is represented by different sensor availability, observation-angle limits, maneuvering capability, energy budget, onboard storage capacity, and resource-consumption coefficients.
Execution heterogeneity is further reflected during decoding, where the same task may correspond to different feasible observation intervals, attitude-transition requirements, energy consumption, storage usage, and load contribution when assigned to different satellites.

Inter-satellite collaboration is considered at the centralized scheduling-decision level.
The scheduler coordinates multiple heterogeneous satellites by assigning tasks to suitable platforms, resolving temporal and attitude-transition conflicts within each satellite sequence, avoiding infeasible or redundant allocations, and balancing workload among satellites.
Explicit inter-satellite communication links, cooperative imaging, and data-relay constraints are beyond the current formulation and are discussed as future extensions.

To preserve these heterogeneous characteristics while maintaining a compact search space, this paper adopts a preference-adjustable weighted-utility model with three normalized components: task gain, energy saving, and load balance.
Candidate schedules are encoded indirectly as task-to-satellite assignment vectors.
A shared decoder then determines feasible observation intervals and execution sequences according to satellite-specific visibility windows, task time windows, temporal conflicts, attitude-transition requirements, payload/resource availability, and onboard resource limits.
The weighted utility is further converted into an equivalent cost by adding penalties for constraint violation and unscheduled feasible tasks.
\begin{table}[!t]
	\centering
	\caption{Main notation}
	\label{tab01}
	{\small
	\setlength{\tabcolsep}{5pt}
	\renewcommand{\arraystretch}{1.06}
	\begin{tabular}{@{}ll@{}}
		\toprule
		Symbol & Description \\
		\midrule
		$\mathcal{S}$ & Set of agile satellites \\
		$\mathcal{T}$ & Set of observation tasks \\
		$N_s$, $N_t$ & Numbers of satellites and tasks \\
		$x_j$ & Assignment variable of task $j$ \\
		$y_{ij}$ & Binary indicator that task $j$ is assigned to satellite $i$ \\
		$\mathcal{W}_{ij}$ & Visibility window set between satellite $i$ and task $j$ \\
		$E_i^{\max}$ & Energy budget of satellite $i$ \\
		$D_i^{\max}$ & Storage capacity of satellite $i$ \\
		$\rho_i^{\mathrm{obs}}$ & Observation energy coefficient of satellite $i$ \\
		$\rho_i^{\mathrm{tr}}$ & Maneuver energy coefficient of satellite $i$ \\
		$b_j$, $f_j$ & Start and finish times of task $j$ \\
		$d_j$ & Required observation duration of task $j$ \\
		$p_j$ & Priority coefficient of task $j$ \\
		$U_g$, $U_e$, $U_b$ & Gain, energy-saving, and load-balance utilities \\
		$U_w$ & Weighted scheduling utility \\
		$CV_{\mathrm{norm}}$ & Normalized constraint violation \\
		$\mathcal{T}^{\mathrm{feas}}$ & Set of tasks observable by at least one satellite \\
		$N_t^{\mathrm{feas}}$ & Number of tasks observable by at least one satellite \\
		\bottomrule
	\end{tabular}
	}
\end{table}

\subsection{Notation}

The main notation used in the proposed scheduling model is summarized in Table~\ref{tab01}.

\subsection{Assignment-Based Scheduling Representation}

An assignment-based indirect encoding is adopted to represent candidate schedules.
This encoding reduces the search dimension by assigning the detailed observation timing and sequencing decisions to the decoder.
Each individual is represented by a task-to-satellite assignment vector:
\begin{equation}
	\mathbf{x} = [x_1, x_2, \ldots, x_{N_t}],\label{eq01}
\end{equation}
where
\begin{equation}
	x_j \in \{0,1,\ldots,N_s\}, \quad \forall j \in \mathcal{T}.\label{eq02}
\end{equation}
Here, $x_j=0$ indicates that task $j$ is not assigned to any satellite, whereas $x_j=i$ indicates that task $j$ is assigned to satellite $i$.
For a compact mathematical description, the corresponding binary assignment indicator is defined as
\begin{equation}
	y_{ij} =
	\begin{cases}
		1, & x_j=i,\\
		0, & \text{otherwise},
	\end{cases}
	\quad i \in \mathcal{S}, \; j \in \mathcal{T}.\label{eq03}
\end{equation}
After decoding, the scheduling state of task $j$ is given by
\begin{equation}
	u_j =
	\begin{cases}
		1, & \text{if task } j \text{ is scheduled},\\
		0, & \text{otherwise}.\label{eq04}
	\end{cases}
\end{equation}

The assignment vector specifies only the candidate satellite for each task.
It does not directly determine the observation start time or execution order.
Given $\mathbf{x}$, the decoder constructs feasible observation intervals and satellite-specific task sequences by checking visibility windows, task time windows, temporal conflicts, attitude-transition feasibility, and onboard resource limits.

\subsection{Utility Components}

The quality of a decoded schedule is evaluated using three normalized utility components: task gain, energy saving, and load balance.
The task-gain utility measures the value of scheduled observations, the energy-saving utility reflects resource efficiency, and the load-balance utility evaluates workload distribution among satellites.

\subsubsection{Task-Gain Utility}

The raw observation gain is defined as
\begin{equation}
	P(\mathbf{x})=\sum_{j\in\mathcal{T}} p_j d_j u_j ,\label{eq05}
\end{equation}
where $p_j$ and $d_j$ denote the priority coefficient and required observation duration of task $j$, respectively.
The ideal maximum gain is given by
\begin{equation}
	P^{\max}=\sum_{j\in\mathcal{T}} p_j d_j .\label{eq06}
\end{equation}
Accordingly, the normalized task-gain utility is defined as
\begin{equation}
	U_g(\mathbf{x})=\frac{P(\mathbf{x})}{P^{\max}+\varepsilon},\label{eq07}
\end{equation}
where $\varepsilon$ is a small positive constant used to avoid division by zero.
A larger $U_g(\mathbf{x})$ indicates that more high-priority observation tasks are successfully scheduled.

\subsubsection{Energy-Saving Utility}

Let $E_i^{\mathrm{obs}}(\mathbf{x})$ and $E_i^{\mathrm{tr}}(\mathbf{x})$ denote the observation and attitude-transition energy consumptions of satellite $i$, respectively.
These quantities are satellite dependent because different platforms may have different observation-energy rates, maneuvering efficiencies, and energy budgets.
The total energy consumption of the schedule is expressed as
\begin{equation}
	E_{\rm tot}(\mathbf{x})=
	\sum_{i\in\mathcal{S}}
	\left(E_i^{\mathrm{obs}}(\mathbf{x})+E_i^{\mathrm{tr}}(\mathbf{x})\right).\label{eq08}
\end{equation}
The normalized energy-saving utility is then defined as
\begin{equation}
	U_e(\mathbf{x})=
	\max\left\{0,1-\frac{E_{\rm tot}(\mathbf{x})}{E_{\rm ref}+\varepsilon}\right\},\label{eq09}
\end{equation}
where $E_{\rm ref}$ is the energy-normalization scale.
A larger $U_e(\mathbf{x})$ corresponds to lower energy consumption and therefore better resource efficiency.

\subsubsection{Load-Balance Utility}

Let $L_i(\mathbf{x})$ denote the normalized scheduling load of satellite $i$, which is calculated from its scheduled observation workload and resource usage relative to its service capability.
The average load over all satellites is defined as
\begin{equation}
	\bar{L}(\mathbf{x})=
	\frac{1}{N_s}\sum_{i\in\mathcal{S}} L_i(\mathbf{x}).\label{eq10}
\end{equation}
The load dispersion is given by
\begin{equation}
	D_L(\mathbf{x})=
	\sum_{i\in\mathcal{S}}
	\left(L_i(\mathbf{x})-\bar{L}(\mathbf{x})\right)^2 .\label{eq11}
\end{equation}
The normalized load-balance utility is defined as
\begin{equation}
	U_b(\mathbf{x})=
	\max\left\{
	0,
	1-
	\frac{D_L(\mathbf{x})}
	{\frac{N_s-1}{N_s}\left(\sum_{i\in\mathcal{S}}L_i(\mathbf{x})\right)^2+\varepsilon}
	\right\}.\label{eq12}
\end{equation}
The denominator normalizes the maximum load dispersion under a fixed total load.
Thus, a larger $U_b(\mathbf{x})$ indicates a more balanced utilization of satellite resources.

\subsection{Preference-Adjustable Weighted Objective}

The three utility components are integrated through a preference weight vector:
\begin{equation}
	\mathbf{w}=(w_g,w_e,w_b),
	w_g+w_e+w_b=1,\quad w_g,w_e,w_b\geq 0,\label{eq13}
\end{equation}
where $w_g$, $w_e$, and $w_b$ denote the weights assigned to task gain, energy saving, and load balance, respectively.
The weighted scheduling utility is then defined as
\begin{equation}
	U_w(\mathbf{x})=
	w_g U_g(\mathbf{x})+
	w_e U_e(\mathbf{x})+
	w_b U_b(\mathbf{x}).\label{eq14}
\end{equation}
The theoretical scheduling objective is to maximize the weighted utility:
\begin{equation}
	\max_{\mathbf{x}} \; U_w(\mathbf{x}).\label{eq15}
\end{equation}

In the experiments, the best weighted utility obtained in one independent run is reported as
\begin{equation}
	U_{\rm best}=
	\max_{\mathbf{x}\in\Omega} U_w(\mathbf{x}).\label{eq16}
\end{equation}
By adjusting $\mathbf{w}$, the same scheduling model can represent different operational preferences, such as gain-prioritized, energy-saving, or load-balanced scheduling.

\subsection{Constraints and Constraint Handling}

A feasible schedule must satisfy task-assignment uniqueness, visibility-window feasibility, observation-duration consistency, temporal-conflict avoidance, attitude-transition feasibility, and onboard resource constraints.
Because the proposed method uses indirect assignment encoding, these constraints are enforced and evaluated during decoding rather than embedded directly in the chromosome representation.

\subsubsection{Task Assignment Constraint}

Each task can be assigned to at most one satellite:
\begin{equation}
	\sum_{i\in\mathcal{S}} y_{ij} \leq 1,
	\quad \forall j\in\mathcal{T}.\label{eq17}
\end{equation}
This constraint is naturally satisfied by the integer assignment encoding in (\ref{eq02}), where each task has only one assignment variable.

\subsubsection{Visibility and Observation-Window Constraint}

Let $\mathcal{W}_{ij}$ denote the set of visibility windows between satellite $i$ and task $j$.
If task $j$ is assigned to satellite $i$, at least one feasible visibility window must exist:
\begin{equation}
	y_{ij} \leq v_{ij},
	\quad \forall i\in\mathcal{S}, \; j\in\mathcal{T},\label{eq18}
\end{equation}
where $v_{ij}=1$ if $\mathcal{W}_{ij}\neq\emptyset$, and $v_{ij}=0$ otherwise.

Let $[a_j,c_j]$ be the allowable time window of task $j$, and let $d_j$ be its required observation duration.
For a scheduled task, there must exist a visibility window $[\ell,u]\in \mathcal{W}_{ij}$ such that
\begin{equation}
	\max(a_j,\ell)\leq b_j,
	\quad
	f_j=b_j+d_j\leq \min(c_j,u).\label{eq19}
\end{equation}
This condition ensures that the decoded observation interval is simultaneously compatible with the task time window and the satellite--task visibility window.

\subsubsection{Temporal and Attitude-Transition Constraint}

For satellite $i$, let $\pi_i=(j_1,j_2,\ldots,j_{m_i})$ denote the decoded execution sequence.
For any two consecutive tasks $j_k$ and $j_{k+1}$ in $\pi_i$, the following condition must hold:
\begin{equation}
	b_{j_{k+1}} \geq f_{j_k}+\tau_{i,j_k,j_{k+1}},
	\quad k=1,\ldots,m_i-1,\label{eq20}
\end{equation}
where $\tau_{i,j_k,j_{k+1}}$ denotes the required attitude-transition time from task $j_k$ to task $j_{k+1}$ on satellite $i$.
This constraint prevents temporal overlap and accounts for the maneuvering time required between consecutive observations.

\subsubsection{Energy and Storage Constraints}

Let $E_i^{\max}$ and $D_i^{\max}$ denote the available energy budget and onboard storage capacity of satellite $i$ over the planning horizon, respectively.
Let $E_i(t)$ and $D_i(t)$ denote the remaining energy and onboard data amount of satellite $i$ at time $t$.
The resource constraints are expressed as
\begin{equation}
	E_i(t)\geq 0,
	\quad
	0\leq D_i(t)\leq D_i^{\max},
	\quad
	\forall i\in\mathcal{S}, \; t\in[0,T_{\rm sim}].
	\label{eq21}
\end{equation}
Because satellites are heterogeneous, the energy budgets, storage capacities, and resource-consumption processes are evaluated separately for each satellite during decoding.

\subsubsection{Constraint-Violation Measure and Equivalent Cost}

Instead of directly discarding infeasible schedules, a normalized constraint-violation measure is introduced to guide the search toward feasible regions.
The total constraint violation is defined as
\begin{equation}
	CV(\mathbf{x})=
	CV_{\rm win}
	+CV_{\rm dur}
	+CV_{\rm tr}
	+CV_{\rm ene}
	+CV_{\rm data},\label{eq22}
\end{equation}
where the five terms denote violations of the time-window, observation-duration, attitude-transition, energy, and data-storage constraints, respectively.
A strictly feasible schedule satisfies $CV(\mathbf{x})=0$, and the corresponding normalized violation is denoted by $CV_{\mathrm{norm}}(\mathbf{x})$.

For compatibility with the minimization interface of the optimizer, the weighted utility is converted into an equivalent cost:
\begin{equation}
	C_w(\mathbf{x})=
	1-U_w(\mathbf{x})
	+\lambda_{\rm cv}CV_{\mathrm{norm}}(\mathbf{x})
	+\lambda_{\rm uns}R_{\rm uns}(\mathbf{x}),\label{eq23}
\end{equation}
where $\lambda_{\rm cv}$ and $\lambda_{\rm uns}$ are penalty coefficients that are constrained to be non-negative.

Let \(\mathcal{T}^{\rm feas}=\{j\in\mathcal{T}|\exists i\in\mathcal{S},\mathcal{W}_{ij}\neq\emptyset\}\) denote the set of tasks observable by at least one satellite, and let \(N_t^{\rm feas}=|\mathcal{T}^{\rm feas}|\). 
Since \(N_t^{\rm feas}>0\) in the tested scenarios, the unscheduled-task ratio is defined as
\begin{equation}
	R_{\rm uns}(\mathbf{x})=
	1-\frac{1}{N_t^{\rm feas}}
	\sum_{j\in\mathcal{T}^{\rm feas}}u_j.
	\label{eq24}
\end{equation}
The optimization problem solved by the proposed algorithm is therefore
\begin{equation}
	\min_{\mathbf{x}} \; C_w(\mathbf{x}).\label{eq25}
\end{equation}
The schedule with the largest weighted utility is finally reported as the scheduling result.

The above formulation also defines the interface between the scheduling model and the optimization algorithm.
The decision variable optimized by the algorithm is the assignment vector \(\mathbf{x}\), whereas observation-window selection, execution sequencing, resource-consumption calculation, and constraint-violation evaluation are completed by the schedule decoder.
For each candidate assignment, the decoder returns the weighted utility \(U_w(\mathbf{x})\), the normalized constraint violation \(CV_{\mathrm{norm}}(\mathbf{x})\), and the equivalent cost \(C_w(\mathbf{x})\).
Therefore, all search operators are evaluated through the same decoder-based scheduling model rather than through an independent surrogate objective.

\section{EPO Framework and RLOSMEA}
\label{sec:framework}

\begin{figure*}[!t]
	\centering
	\includegraphics[width=\textwidth]{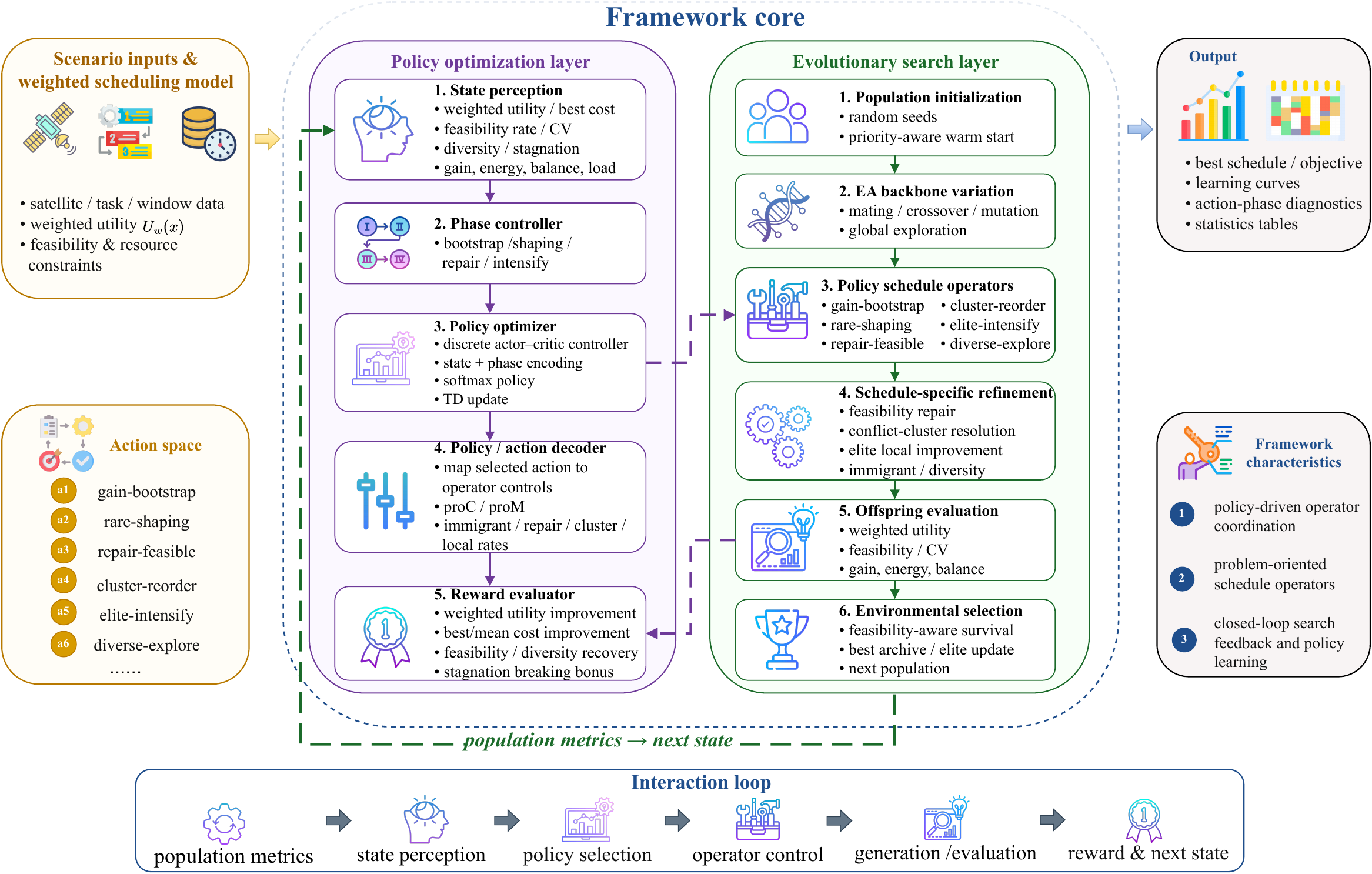}
	\caption{Evolutionary policy optimization framework for heterogeneous AEOS scheduling.}
	\label{fig02}
\end{figure*}

\subsection{Framework Motivation and Overview}

The proposed EPO framework is built on the assignment-based scheduling model defined in Section~\ref{sec:model}.
The assignment vector \(\mathbf{x}\) serves as the individual representation in the evolutionary population, where each element specifies whether a task is unscheduled or assigned to a candidate satellite.
Given an assignment vector, the shared decoder determines satellite-specific observation intervals, execution sequences, resource consumption, and constraint violations according to the visibility, temporal, attitude-transition, energy, storage, and load-related constraints introduced above.
Thus, RLOSMEA searches over the assignment space defined by the scheduling model, while all candidate solutions are evaluated using the same decoder-based equivalent cost \(C_w(\mathbf{x})\).

As shown in Fig.~\ref{fig02}, the framework integrates population-based evolutionary search with online actor--critic policy control.
The evolutionary module generates and updates candidate assignments, whereas the policy module observes the population state and selects a high-level operator mode for offspring generation and schedule modification.
The policy acts as an adaptive operator controller rather than a direct schedule constructor, which keeps the RL action space compact and preserves the global search capability of evolutionary optimization.

RLOSMEA is developed as an AEOS-oriented implementation of the proposed framework.
It combines a memetic evolutionary backbone with online operator-mode selection and schedule-specific modification operators, including warm start, feasibility repair, conflict resolution, local improvement, elite intensification, immigrant injection, and diversity rescue.
All schedule-oriented modifications are applied before decoding, so each offspring is evaluated once by the shared decoder under the fixed function-evaluation budget.

\subsection{The MDP Formulation}

The adaptive operator-mode selection process is formulated as an online Markov decision process (MDP).
To distinguish it from the AEOS scheduling model in Section~\ref{sec:model}, the MDP for operator selection is defined as
\begin{equation}
	\mathcal{M}_{\rm RL}=(\mathcal{X},\mathcal{U},\mathcal{K},r,\gamma),\label{eq26}
\end{equation}
where $\mathcal{X}$ is the population-state space, $\mathcal{U}$ is the operator-mode action space, $\mathcal{K}(\xi_{t+1}|\xi_t,\alpha_t)$ denotes the transition kernel induced by action $\alpha_t$, $r(\xi_t,\alpha_t,\xi_{t+1})$ is the reward function, and $\gamma$ is the discount factor.

At generation $t$, the actor--critic controller observes the population state $\xi_t\in\mathcal{X}$ and selects an operator-mode action $\alpha_t\in\mathcal{U}$.
After offspring generation, schedule modification, decoding evaluation, and environmental selection, the controller receives the immediate reward
\begin{equation}
	\varrho_t=r(\xi_t,\alpha_t,\xi_{t+1}).\label{eq27}
\end{equation}

Learning is performed online in each independent run without using an offline training dataset.
The controller is updated generation by generation using the transition tuple
\begin{equation}
	(\xi_t,\alpha_t,\varrho_t,\xi_{t+1}).\label{eq28}
\end{equation}

This MDP is defined for online operator-mode selection rather than direct schedule construction.
The transition from \(\xi_t\) to \(\xi_{t+1}\) is induced by applying the selected operator mode to the current population, generating and modifying offspring assignment vectors, decoding the resulting schedules, evaluating \(C_w(\mathbf{x})\), and performing environmental selection.
Thus, the RL controller interacts with the scheduling problem through the population and the shared decoder.
The state, action, and reward are all constructed from decoder-evaluated scheduling information, while the final optimization target remains the equivalent cost defined in Section~\ref{sec:model}.

\subsection{Population-State Representation}

The population state provides compact contextual information for online operator-mode selection.
Instead of encoding a complete schedule, it summarizes the current search condition from four aspects: search progress, scheduling quality, feasibility, and population structure.
The state vector observed by the actor--critic controller at generation $t$ is written as
\begin{equation}
	\xi_t=
	\left[
	\xi_t^{p},
	\xi_t^{u},
	\xi_t^{f},
	\xi_t^{s}
	\right]^T,\label{eq29}
\end{equation}
where $\xi_t^{p}$, $\xi_t^{u}$, $\xi_t^{f}$, and $\xi_t^{s}$ denote the progress, utility, feasibility, and population-structure features, respectively.
These features are used only to guide operator-mode selection and do not modify the scheduling objective defined in Section~\ref{sec:model}.

The state features are derived from decoded schedules rather than raw assignment vectors alone.
Utility-related features are computed from \(U_w(\mathbf{x})\) and its component utilities, feasibility-related features are obtained from \(CV_{\mathrm{norm}}(\mathbf{x})\) and feasible-task ratios, and population-structure features describe diversity and fitness dispersion among decoded individuals.
Additional diagnostics, such as priority capture, energy consumption, load balance, stagnation, and offspring success rate, summarize the current search condition under heterogeneous satellite constraints.
These features guide operator-mode selection without modifying the scheduling objective defined in Section~\ref{sec:model}.

\subsection{Search-Phase Descriptor}

A search-phase descriptor is introduced to provide stage-dependent information for policy evaluation.
The phase set is defined as
\begin{equation}
	\mathcal{H}=
	\{\mathrm{bootstrap},\mathrm{shaping},\mathrm{repair},\mathrm{intensify}\}.\label{eq30}
\end{equation}
The current phase is determined from the population state and recent search diagnostics, including progress, feasibility, constraint violation, diversity, priority capture, load balance, stagnation, and offspring success rate.
The phase descriptor is encoded as a one-hot vector $\psi_t$ and concatenated with the population state before action selection.

The bootstrap phase promotes the rapid construction of promising assignment structures.
The shaping phase emphasizes rare-task enhancement and structural adjustment.
The repair phase is activated when feasibility is poor or constraint violation is high.
The intensification phase is used when the population enters a promising region and local refinement becomes more beneficial.

The phase descriptor is therefore a scheduling-aware abstraction of the current search stage.
It helps the policy distinguish whether the population mainly requires assignment construction, rare-task preservation, feasibility recovery, or local intensification, rather than applying a fixed operator schedule throughout the search.

A rule-based recovery mode is retained as a safeguard against severe stagnation or diversity collapse.
When recovery is triggered, the selected action $\alpha_t$ can be replaced by an exploration-oriented or restructuring action in $\mathcal{U}$.
Such recovery-forced actions are not used to update the actor policy, thereby preventing externally forced decisions from biasing policy learning.

\subsection{Operator-Mode Action Space}

The actor policy selects a high-level operator-mode action rather than a direct scheduling decision.
The action space is defined as
\begin{equation}
	\mathcal{U}=\{\alpha_1,\alpha_2,\alpha_3,\alpha_4,\alpha_5,\alpha_6\}.\label{eq31}
\end{equation}
The six operator modes are summarized in Table~\ref{tab02}.
Each action represents a coordinated search mode for modifying offspring assignment vectors.
After action $\alpha_t$ is selected, it is decoded into the operator-control vector

\begin{equation}
	\boldsymbol{\Theta}_t=(p_{c,t},p_{m,t},\boldsymbol{\eta}_t),\label{eq32}
\end{equation}
where $p_{c,t}$ and $p_{m,t}$ denote the crossover and mutation parameters, respectively, and $\boldsymbol{\eta}_t$ denotes the vector of schedule-specific operator rates.
These rates control the activation intensities of warm start, feasibility repair, conflict resolution, local improvement, elite intensification, immigrant injection, rare-task shaping, and diversity rescue.

\begin{table}[!t]
	\centering
	\caption{Operator-mode actions}
	\label{tab02}
	{\small
		\setlength{\tabcolsep}{5pt}
		\renewcommand{\arraystretch}{1.06}
		\begin{tabular}{@{}c l l@{}}
			\toprule
			Action & Mode & Main Function \\
			\midrule
			$\alpha_1$ & gain-bootstrap & Gain construction \\
			$\alpha_2$ & rare-shaping & Rare-task enhancement \\
			$\alpha_3$ & repair-feasible & Feasibility recovery \\
			$\alpha_4$ & cluster-reorder & Conflict resolution \\
			$\alpha_5$ & elite-intensify & Elite exploitation \\
			$\alpha_6$ & diverse-explore & Diversity recovery \\
			\bottomrule
		\end{tabular}
	}
\end{table}

The action decoder follows a sparse-control principle.
For each selected action, one dominant problem-oriented operator is activated, whereas the remaining operators are retained as weak background safeguards.
This design improves the interpretability of the selected action and provides a clearer reward signal for policy learning.

\subsection{Actor--Critic Policy Optimization}

Let $\pi_{\theta}(\alpha|\xi_t,\psi_t)$ denote the actor policy parameterized by $\theta$.
The input feature vector for the actor--critic controller is constructed as
\begin{equation}
	\boldsymbol{\phi}_t=[\xi_t;\psi_t;1],\label{eq33}
\end{equation}
where the last entry is a bias term.
The actor computes the action logits as
\begin{equation}
	\mathbf{o}_t=\mathbf{W}\boldsymbol{\phi}_t+\mathbf{b}_{\psi_t},\label{eq34}
\end{equation}
where $\mathbf{W}$ is the actor parameter matrix, and $\mathbf{b}_{\psi_t}$ is the phase-dependent bias vector.

A temperature-controlled softmax function is first used to obtain the preliminary policy:
\begin{equation}
	\pi_{\theta}^{0}(\alpha|\xi_t,\psi_t)
	=
	\frac{\exp(o_{\alpha,t}/\sigma)}
	{\sum_{\alpha'\in\mathcal{U}}\exp(o_{\alpha',t}/\sigma)},\label{eq35}
\end{equation}
where $\sigma$ is the temperature parameter.
The phase prior $q_{\psi_t}(\alpha)$ is then incorporated as
\begin{equation}
	\pi_{\theta}(\alpha|\xi_t,\psi_t)
	=
	\frac{
		\pi_{\theta}^{0}(\alpha|\xi_t,\psi_t)q_{\psi_t}(\alpha)
	}
	{
		\sum_{\alpha'\in\mathcal{U}}
		\pi_{\theta}^{0}(\alpha'|\xi_t,\psi_t)q_{\psi_t}(\alpha')
	}.\label{eq36}
\end{equation}
To maintain exploration, an $\epsilon$-mixed policy is adopted:
\begin{equation}
	\tilde{\pi}_{\theta}(\alpha|\xi_t,\psi_t)
	=
	(1-\epsilon)\pi_{\theta}(\alpha|\xi_t,\psi_t)
	+
	\epsilon\frac{1}{|\mathcal{U}|}.\label{eq37}
\end{equation}

The critic estimates the state value using a linear approximation:
\begin{equation}
	V_{\omega}(\xi_t,\psi_t)=\mathbf{v}^{T}\boldsymbol{\phi}_t,\label{eq38}
\end{equation}
where $\mathbf{v}$ is the critic parameter vector.
After action $\alpha_t$ is applied and the next state is obtained, the temporal-difference error is computed as
\begin{equation}
	\delta_t=
	\varrho_t+
	\gamma V_{\omega}(\xi_{t+1},\psi_{t+1})
	-
	V_{\omega}(\xi_t,\psi_t).\label{eq39}
\end{equation}
For numerical stability, $\delta_t$ is clipped within a finite interval in the implementation.

For a policy-sampled action, the actor and critic are updated as
\begin{equation}
	\mathbf{W}
	\leftarrow
	\mathbf{W}
	+
	\alpha_a\delta_t
	(\mathbf{e}_{\alpha_t}-\tilde{\boldsymbol{\pi}}_t)
	\boldsymbol{\phi}_t^{T},\label{eq40}
\end{equation}
\begin{equation}
	\mathbf{v}
	\leftarrow
	\mathbf{v}
	+
	\alpha_c\delta_t\boldsymbol{\phi}_t,\label{eq41}
\end{equation}
where $\alpha_a$ and $\alpha_c$ are the actor and critic learning rates, respectively, $\mathbf{e}_{\alpha_t}$ is the one-hot vector of the selected action, and $\tilde{\boldsymbol{\pi}}_t$ is the mixed action-probability vector.
If action $\alpha_t$ is forced by the recovery mode, the actor update is skipped to avoid biasing the learned policy with safeguard decisions.

\subsection{Reward Function}

The reward function is designed to align online operator-mode selection with the weighted scheduling objective. It consists of four terms: weighted-utility improvement, best-cost improvement, mean-cost improvement, and offspring success rate. Feasibility, diversity, and stagnation are included in the state and phase descriptors rather than being directly added as separate reward terms. This design keeps the reward compact and avoids excessive reward shaping.

Let $P_t$ and $P_{t+1}$ denote the populations before and after the offspring generation, decoder evaluation, and environmental selection induced by the selected operator mode, respectively. For $k\in\{t,t+1\}$, the mean weighted utility of population $P_k$ is calculated as
\begin{equation}
	\bar{U}_k=
	\frac{1}{|P_k|}
	\sum_{\mathbf{x}\in P_k}U_w(\mathbf{x}).
\end{equation}
Accordingly, the weighted-utility improvement is defined as
\begin{equation}
	\Delta U_t=\bar{U}_{t+1}-\bar{U}_t.
\end{equation}

Since the optimizer minimizes the equivalent cost $C_w(\mathbf{x})$, the best and mean equivalent costs of population $P_k$ are calculated as
\begin{equation}
	C_k^{\rm best}=
	\min_{\mathbf{x}\in P_k}C_w(\mathbf{x}),
\end{equation}
and
\begin{equation}
	C_k^{\rm mean}=
	\frac{1}{|P_k|}
	\sum_{\mathbf{x}\in P_k}C_w(\mathbf{x}),
\end{equation}
respectively. The corresponding best-cost and mean-cost improvements are then defined as
\begin{equation}
	\Delta C_t^{\rm best}
	=
	\frac{C_t^{\rm best}-C_{t+1}^{\rm best}}
	{\max\left(|C_t^{\rm best}|,\varepsilon_c\right)},
\end{equation}
and
\begin{equation}
	\Delta C_t^{\rm mean}
	=
	\frac{C_t^{\rm mean}-C_{t+1}^{\rm mean}}
	{\max\left(|C_t^{\rm mean}|,\varepsilon_c\right)},
\end{equation}
respectively, where $\varepsilon_c$ is a small positive constant used to avoid numerical instability when the reference cost approaches zero.

The offspring success rate is defined as

\begin{equation}
\begin{aligned}
	Q_t^{\rm succ}
	=&\frac{1}{|Q_t|}
	\sum_{\mathbf{x}_i\in Q_t}
	\mathbb{I}\Big(C_w(\mathbf{x}_i)
	<\operatorname{median}\{C_w(\mathbf{x})|\mathbf{x}\in P_t\}\Big).
\end{aligned}\label{eq45}
\end{equation}
where $P_t$ and $Q_t$ denote the parent and offspring populations, respectively, and $\mathbb{I}(\cdot)$ is the indicator function.

The immediate reward is then written as
\begin{equation}
	\varrho_t=
	{\rm clip}
	\left(
	\sum_{j=1}^{4}\lambda_j\Gamma_j(t),
	-1,1
	\right),\label{eq46}
\end{equation}
where
\begin{equation}
	\Gamma_1(t)=\tanh\left(\frac{\Delta U_t}{s_1}\right),
	\Gamma_2(t)=\tanh\left(\frac{\Delta C_t^{\rm best}}{s_2}\right),\label{eq47}
\end{equation}
\begin{equation}
	\Gamma_3(t)=\tanh\left(\frac{\Delta C_t^{\rm mean}}{s_3}\right),
	\Gamma_4(t)=\tanh\left(\frac{Q_t^{\rm succ}-q_0}{s_4}\right).\label{eq48}
\end{equation}

The reward is not an independent learning objective separated from the scheduling model.
It is a compact feedback signal derived from the same decoder-based evaluation used for optimization.
The utility-improvement term is computed from \(U_w(\mathbf{x})\), whereas the best-cost and mean-cost improvement terms are computed from \(C_w(\mathbf{x})\).
The offspring-success term measures whether the selected operator mode generates competitive decoded schedules.
Feasibility, diversity, priority capture, energy consumption, and load-balance information are represented in the population state and phase descriptor rather than added as separate reward terms, thereby avoiding excessive reward shaping and credit-assignment noise.

\subsection{Policy-Controlled Evolutionary Search}

\begin{algorithm}[!t]
\caption{RLOSMEA Under the EPO Framework}
\label{alg:RLOSMEA}
\begin{algorithmic}[1]
\REQUIRE Scheduling instance \(I\), shared decoder \(\mathcal{D}\), population size \(N\), maximum function evaluations \(N_{\mathrm{FE}}^{\max}\), preference weight vector \(w\), and fixed RLOSMEA parameters.
\ENSURE Best scheduling solution \(x^{*}\).

\STATE Prepare problem-guidance information \(G\), including task priorities, task durations, feasible satellite sets, visibility-window statistics, window slack, and satellite resource indicators.
\STATE Initialize population \(P_0\), decode and evaluate all individuals using \(\mathcal{D}\), and extract the initial population state \(\xi_0\).
\STATE Initialize the incumbent best solution \(x^{*}\), actor parameter \(W\), critic parameter \(v\), exploration parameters \((\epsilon,\sigma)\), and search diagnostics.
\STATE Set \(N_{\mathrm{FE}}\leftarrow |P_0|\) and \(t\leftarrow 0\).

\WHILE{\(N_{\mathrm{FE}}<N_{\mathrm{FE}}^{\max}\)}
    \STATE Determine the search-phase descriptor \(\psi_t\) from the current population state and search diagnostics.
    \STATE Construct the policy input \(\phi_t=[\xi_t;\psi_t;1]\), compute the phase-aware mixed policy, and sample an operator-mode action \(\alpha_t\in\mathcal{U}\).
    \STATE Set the recovery flag \(r_t\leftarrow 0\).
    \IF{severe stagnation or diversity collapse is detected}
        \STATE Replace \(\alpha_t\) with a recovery-oriented action and set \(r_t\leftarrow 1\).
    \ENDIF
    \STATE Decode \(\alpha_t\) into the operator-control vector \(\Theta_t=(p_{c,t},p_{m,t},\eta_t)\).
    \STATE Generate offspring \(Q_t\) through evolutionary variation and apply schedule-specific modifications according to \(\Theta_t\).
    \STATE Decode and evaluate each offspring \(x\in Q_t\) using \(\mathcal{D}\) to obtain \(U_w(x)\), \(CV_{\mathrm{norm}}(x)\), and \(C_w(x)\).
    \STATE Update \(N_{\mathrm{FE}}\leftarrow N_{\mathrm{FE}}+|Q_t|\).
    \STATE Merge \(P_t\) and \(Q_t\), and perform feasibility-aware environmental selection to obtain \(P_{t+1}\).
    \STATE Update the incumbent best solution \(x^{*}\) according to the decoded weighted utility.
    \STATE Extract the next population state \(\xi_{t+1}\), determine \(\psi_{t+1}\), and compute the immediate reward \(\varrho_t\).
    \STATE Compute the temporal-difference error \(\delta_t\) and update the critic parameter \(v\).
    \IF{\(r_t=0\)}
        \STATE Update the actor parameter \(W\).
    \ENDIF
    \STATE Update search diagnostics and set \(t\leftarrow t+1\).
\ENDWHILE

\STATE \textbf{return} \(x^{*}\).
\end{algorithmic}
\end{algorithm}

Algorithm~\ref{alg:RLOSMEA} summarizes the implementation flow of RLOSMEA under the EPO framework. 
In each generation, the decoded population is mapped to a population state, and the actor--critic controller selects an operator-mode action according to the current state and search-phase descriptor. 
The selected action is decoded into the operator-control vector \(\boldsymbol{\Theta}_t\), which guides offspring generation and schedule-specific modifications, including feasibility repair, conflict resolution, local improvement, elite intensification, immigrant injection, and diversity rescue. 
All modifications are completed before decoder-based evaluation, so each offspring is evaluated only once by the shared schedule decoder under the fixed function-evaluation budget. 
The evaluated offspring and parent population are then processed by feasibility-aware environmental selection to form \(\mathcal{P}_{t+1}\), from which the next state, reward, and actor--critic update are obtained. 
This procedure couples policy learning, evolutionary variation, schedule refinement, decoder-based evaluation, and environmental selection within each generation.

\subsection{Complexity Analysis}

Let \(N_t\), \(N\), and \(N_{\mathrm{FE}}^{\max}\) denote the number of tasks, the population size, and the maximum number of function evaluations, respectively. Let \(\bar{\mathcal{W}}\) be the average number of candidate visibility windows checked for a task-satellite assignment. In RLOSMEA, the dominant computational cost comes from decoding and evaluating candidate assignment vectors. For one evaluated individual, the schedule decoder needs to construct satellite-specific task sequences, check feasible visibility windows, and evaluate temporal-transition and resource constraints. Thus, the decoding cost can be approximated as

\begin{equation}
	C_{\mathrm{dec}}=O(N_t\log N_t+N_t\bar{\mathcal{W}}),
\end{equation}
where the sorting and sequence construction operations contribute to the \(N_t\log N_t\) term, and the visibility-window and feasibility checks contribute to the \(N_t\bar{\mathcal{W}}\) term.

Under the fixed function-evaluation budget, each offspring is decoded and evaluated once. Therefore, the overall time complexity of RLOSMEA is approximately

\begin{equation}
	O\left(N_{\mathrm{FE}}^{\max} C_{\mathrm{dec}}\right)
	=
	O\left(N_{\mathrm{FE}}^{\max}(N_t\log N_t+N_t\bar{\mathcal{W}})\right).
\end{equation}

The actor--critic update and operator-mode selection involve only a compact state-action representation and a small number of operator modes.
Their overhead is therefore much smaller than the repeated decoder-based schedule evaluation.

The main memory consumption comes from the population assignment vectors and the preprocessed visibility windows. Therefore, the space complexity of RLOSMEA is

\begin{equation}
	O(NN_t+|\mathcal{W}|),
\end{equation}
where \(|\mathcal{W}|\) denotes the total number of stored visibility windows. The actor--critic parameters introduce only a small additional memory cost and do not change the dominant space complexity.

\subsection{Methodological Discussion}

The proposed EPO framework separates the roles of schedule evaluation, evolutionary search, and policy learning.
The evolutionary component searches over assignment vectors, and the shared decoder evaluates feasibility, resource consumption, weighted utility, and equivalent cost according to the scheduling model in Section~\ref{sec:model}.
The policy component does not construct schedules directly; instead, it selects high-level operator modes that regulate offspring generation and schedule-specific modification.
Thus, reinforcement learning controls the search behavior while leaving physical feasibility and objective evaluation to the decoder.

In RLOSMEA, the population-state representation, phase descriptor, and compact reward jointly enable the controller to switch among construction, repair, exploitation, and exploration under a fixed function-evaluation budget.
The optimizer operates on the equivalent cost, whereas scheduling performance is reported using the weighted utility.
This separation keeps the optimization interface compatible with minimization-based evolutionary search while retaining an interpretable utility-based evaluation of scheduling quality.

\section{Simulation Experiments and Analysis}
\label{sec:experiments}

\subsection{Experimental Settings}

To evaluate effectiveness and scalability, six heterogeneous AEOS scheduling scenarios, denoted as scenario\_01--scenario\_06, are constructed. As listed in Table~\ref{tab03}, the number of candidate tasks increases from 100 to 350, and the number of agile satellites increases from 5 to 12. These scenarios are designed to cover different scheduling scales rather than to impose a linear growth pattern on scheduling difficulty. In heterogeneous AEOS scheduling, the effective difficulty is jointly determined by task scale, satellite--task visibility relations, temporal-window overlap, attitude-transition constraints, onboard resources, and load-distribution requirements.

\begin{table}[!t]
	\centering
	\caption{Heterogeneous AEOS scenarios}
	\label{tab03}
	{\small
		\setlength{\tabcolsep}{8pt}
		\renewcommand{\arraystretch}{1.06}
		\begin{tabular}{@{}lcc@{}}
			\toprule
			Scenario & \(N_t\) & \(N_s\) \\
			\midrule
			scenario\_01 & 100 & 5  \\
			scenario\_02 & 150 & 5  \\
			scenario\_03 & 200 & 7  \\
			scenario\_04 & 250 & 9  \\
			scenario\_05 & 300 & 10 \\
			scenario\_06 & 350 & 12 \\
			\bottomrule
		\end{tabular}
	}
\end{table}

Each task is associated with a ground target, a priority value, an observation duration, and a task time window. Satellite--task visibility windows are generated before optimization by considering orbital motion, target motion, sensor-access constraints, and Earth rotation. The target set contains both static and moving targets. Specifically, $15\%$ of the targets are randomly selected as static targets, while the speed of each remaining moving target is independently sampled from $U(0,45)$ m/s. Each moving target is initialized with a random heading in $[0^\circ,360^\circ)$ and assigned a direction-keeping probability $p_i^{\rm keep}\sim U(0.70,0.95)$. At each $1$-s update step, the heading is retained with probability $p_i^{\rm keep}$; otherwise, a heading perturbation is sampled from $U(-30^\circ,30^\circ)$. The target position is then propagated over the spherical Earth according to the updated heading and target-specific speed. These trajectories are used during preprocessing to generate time-dependent satellite--task access windows for schedule decoding.

The six scenarios therefore contain $5$, $5$, $7$, $9$, $10$, and $12$ agile satellites, respectively. Each satellite follows a near-circular low-Earth orbit. For satellite $i$, the orbital altitude and inclination are defined as $h_i=450+50\operatorname{mod}(i-1,5)$ km and $I_i=30+10\operatorname{mod}(i-1,6)$ deg, respectively, yielding five altitude levels of $\{450,500,550,600,650\}$ km and six inclination levels of $\{30,40,50,60,70,80\}$ deg. The initial longitudes and orbital phases are evenly spaced over $360^\circ$ within each scenario to diversify ground-track coverage. The maximum observation angle of satellite $i$ is defined as $30+5\operatorname{mod}(i-1,4)$ deg, corresponding to four observation-angle classes of $30^\circ$, $35^\circ$, $40^\circ$, and $45^\circ$. Together, these differences in orbital altitude, inclination, initial geometry, and observation angle generate satellite-dependent ground tracks, visibility cones, and access-window structures, thereby instantiating orbital heterogeneity. Payload and resource heterogeneity is further introduced through satellite-specific energy budgets, storage capacities, power-consumption parameters, and data rates, which affect the feasibility and equivalent cost of decoded schedules. The simulation horizon is $86,400$ s, corresponding to a one-day planning horizon. The preprocessing time step is set to $1$ s for satellite--task visibility-window generation, which provides sufficient temporal resolution to reduce visibility-window boundary errors while keeping the preprocessing cost manageable. Task priorities are randomly generated from the discrete set $\{1,2,3,4,5\}$.

All compared algorithms use the same scenario files, visibility windows, decoder, objective function, constraint-evaluation rules, stopping criterion, and number of independent runs. The compared methods include the memetic evolutionary algorithm (MemeticEA)~\cite{39}, adaptive large neighborhood search (ALNS)~\cite{19}, adaptive exploration state-space particle swarm optimization (AESSPSO)~\cite{40}, grey wolf optimizer (GWO)~\cite{41}, reinforcement learning-based memetic algorithm (RLMA)~\cite{42}, and the proposed RLOSMEA. RLMA is included as a recent RL-assisted memetic baseline. In RLMA, reinforcement learning mainly selects local-search operators from an ensemble of neighborhood structures, whereas RLOSMEA uses an online actor--critic controller to select high-level AEOS-oriented operator modes during the evolutionary search process. These modes coordinate gain construction, rare-task shaping, feasibility repair, conflict resolution, elite intensification, and diversity exploration.

\begin{figure*}[!t]
	\centering
	\captionsetup[subfigure]{font=scriptsize}
	\subfloat[scenario\_01]{
		\includegraphics[width=0.48\textwidth]{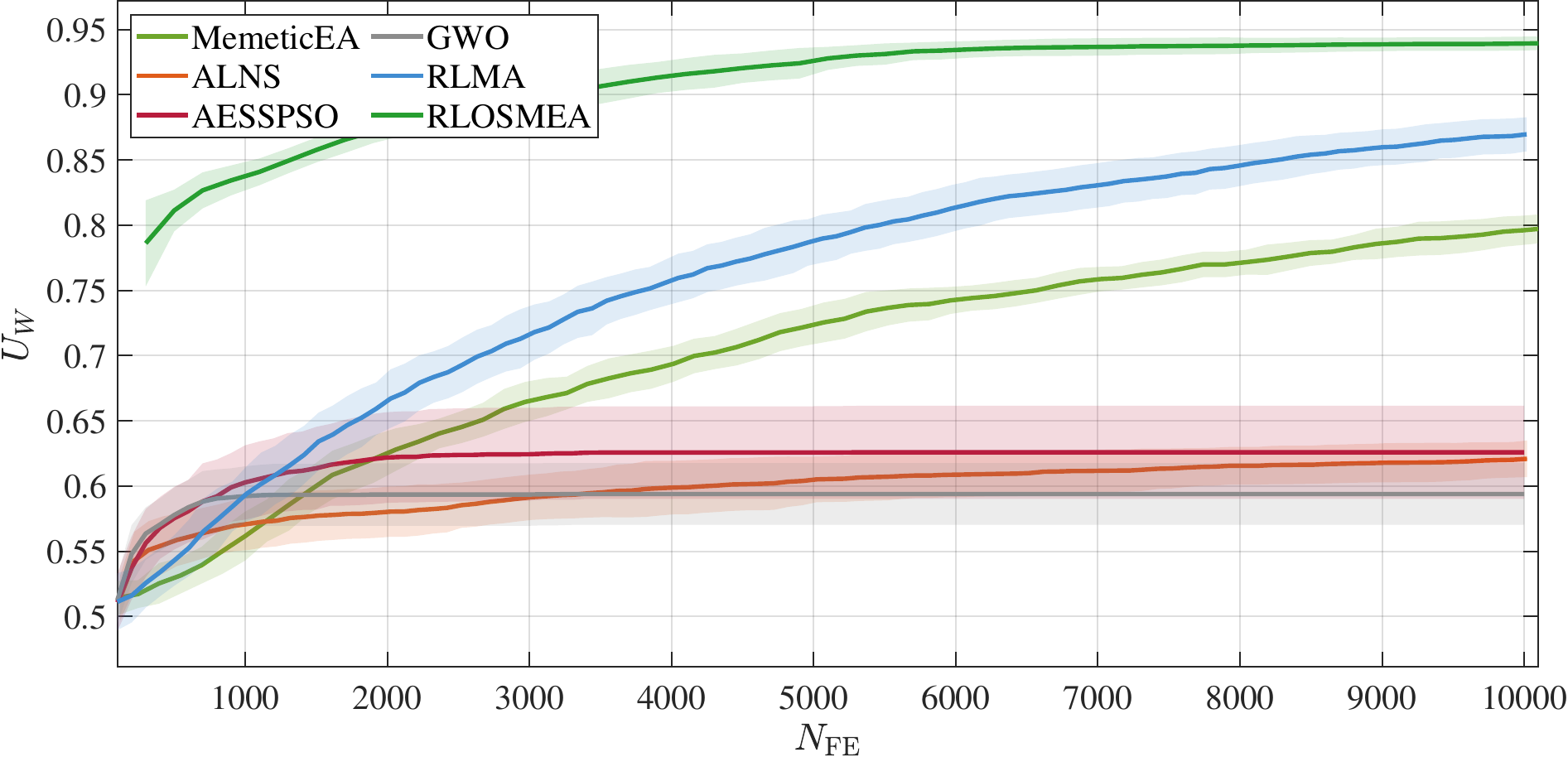}
	}
	\hfil
	\subfloat[scenario\_02]{
		\includegraphics[width=0.48\textwidth]{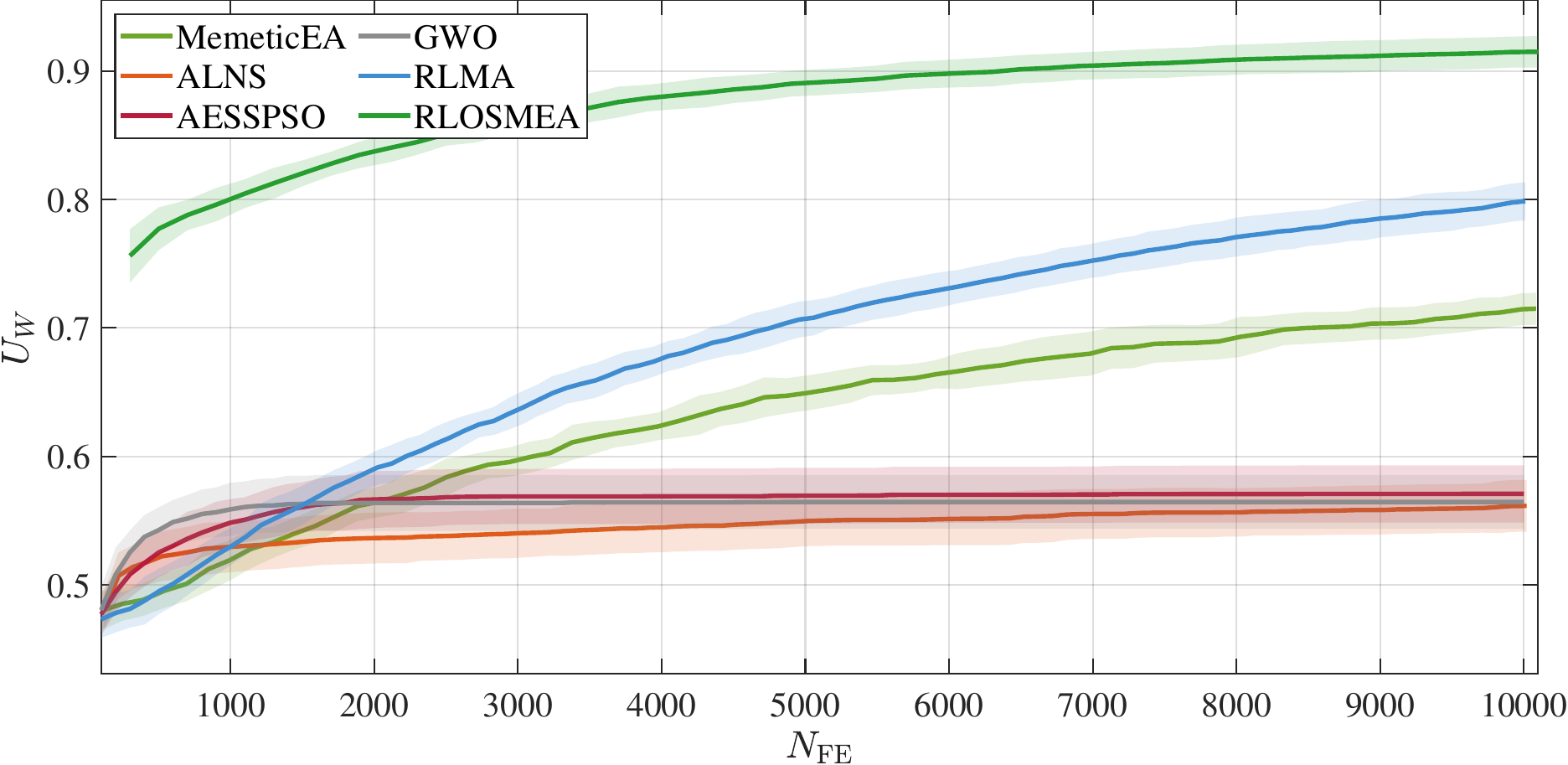}
	}
	\caption{Weighted-utility convergence curves in two representative heterogeneous scenarios. The solid curves show the mean results over 30 independent runs, and the shaded bands indicate the corresponding standard deviations.}
	\label{fig03}
\end{figure*}

For RLOSMEA, all actor--critic and operator-control parameters are fixed across all scenarios and are not tuned separately for individual instances. Specifically, $(\mathrm{actorLR},\mathrm{criticLR})=(0.035,0.08)$ and $(\epsilon,\sigma)=(0.10,0.82)$, where $\epsilon$ and $\sigma$ denote the exploration rate and softmax temperature, respectively. The operator-control base rates are $(\mathrm{immigrantBase},\mathrm{repairBase})=(0.10,0.10)$ and $(\mathrm{clusterBase},\mathrm{localBase})=(0.10,0.07)$. The reward coefficients for weighted-utility improvement, best-cost improvement, mean-cost improvement, and offspring success rate are fixed as $1.45$, $0.90$, $0.35$, and $0.18$, respectively. Each algorithm is independently run 30 times for each scenario with a budget of 10,000 function evaluations. All experiments are conducted on a 64-bit Windows platform equipped with an Intel(R) Core(TM) Ultra 9 285H CPU at 2.90 GHz and 64 GB RAM.

Before comparative analysis, all candidate solutions are decoded into executable satellite-specific schedules, including satellite-specific observation intervals and execution sequences. The decoded schedules are checked under visibility, observation-duration, temporal-conflict, attitude-transition, energy, storage, and load-related constraints. Therefore, all reported performance results are obtained from decoder-validated executable schedules, and the resulting weighted utility and equivalent cost are used for algorithmic comparison.

\subsection{Overall Comparison on Heterogeneous Scenarios}

\subsubsection{Convergence Behavior}

Fig.~\ref{fig03} shows the weighted-utility convergence curves in two representative scenarios. The solid curves denote the mean values over 30 independent runs, and the shaded bands denote the corresponding standard deviations. RLOSMEA improves more rapidly and converges to higher weighted utility than the compared algorithms, indicating that online operator-mode control effectively coordinates exploration, feasibility recovery, and refinement under heterogeneous scheduling constraints.

\begin{figure}[!t]
	\centering
	\captionsetup[subfigure]{font=scriptsize}
	\subfloat[scenario\_01]{
		\includegraphics[width=0.99\columnwidth]{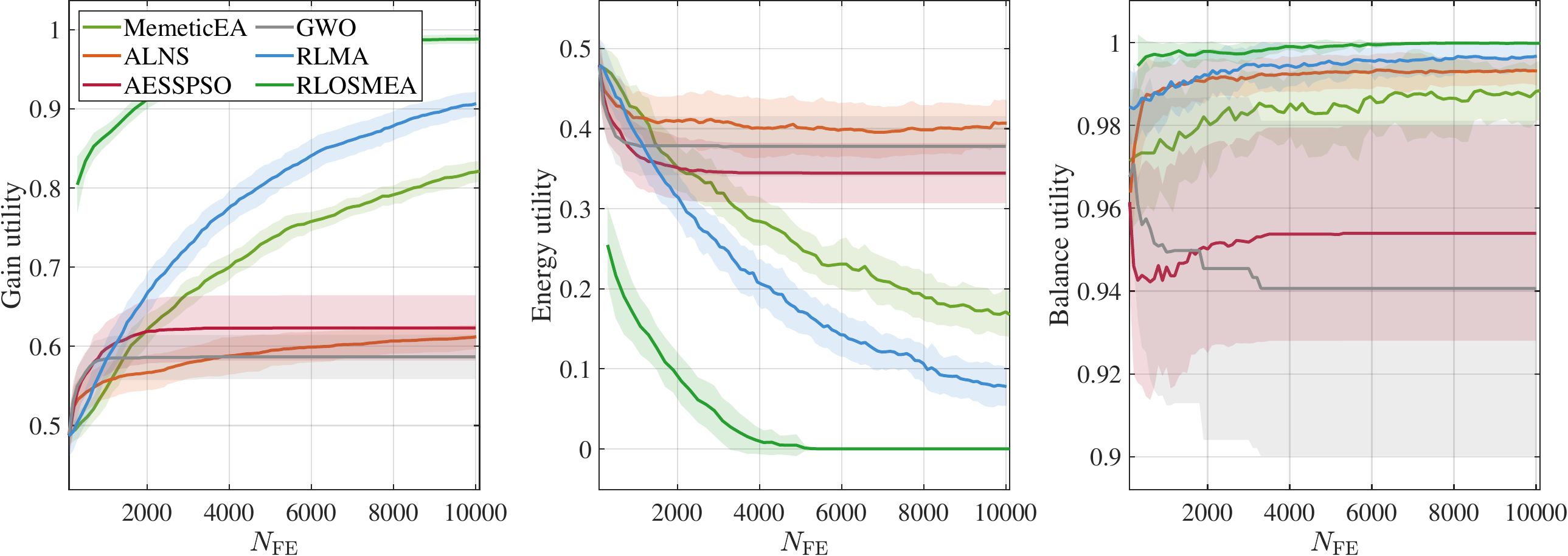}
	}

	\subfloat[scenario\_02]{
		\includegraphics[width=0.99\columnwidth]{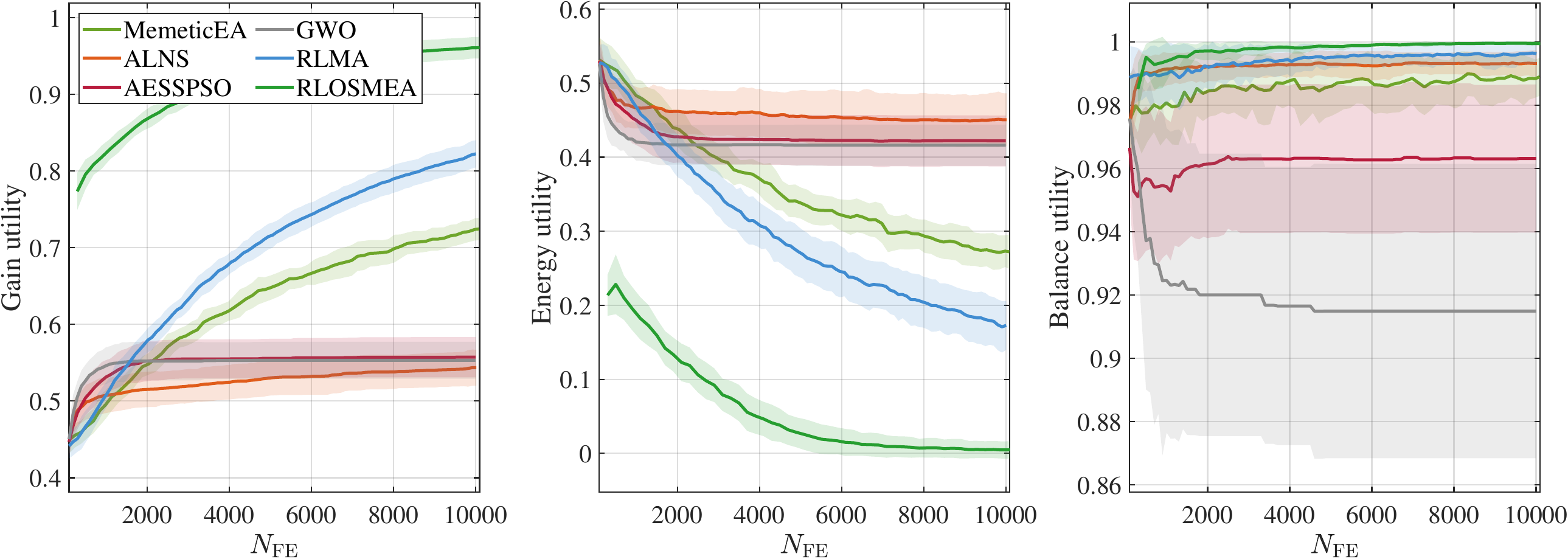}
	}
	\caption{Component-wise utility evolution in representative heterogeneous scenarios. The solid curves show the mean results over 30 independent runs, and the shaded bands indicate the corresponding standard deviations.}
	\label{fig04}
\end{figure}

\begin{table*}[!t]
\centering
\caption{Comparison of six algorithms across six scenarios under random task priorities. Each value reports the mean $\pm$ standard deviation of the weighted utility over 30 independent runs.}
\label{tab04}
\renewcommand{\arraystretch}{2}
\begin{tabular}{@{}l c c c c c c@{}}
\toprule
Algorithm & scenario\_01 & scenario\_02 & scenario\_03 & scenario\_04 & scenario\_05 & scenario\_06 \\
\midrule
MemeticEA & 0.7971$_{\pm0.0111}$ & 0.7149$_{\pm0.0125}$ & 0.7407$_{\pm0.0092}$ & 0.7579$_{\pm0.0087}$ & 0.7395$_{\pm0.0082}$ & 0.7308$_{\pm0.0061}$ \\
ALNS      & 0.6209$_{\pm0.0140}$ & 0.5615$_{\pm0.0201}$ & 0.5912$_{\pm0.0178}$ & 0.6434$_{\pm0.0123}$ & 0.6406$_{\pm0.0141}$ & 0.6431$_{\pm0.0160}$ \\
AESSPSO   & 0.6258$_{\pm0.0357}$ & 0.5708$_{\pm0.0222}$ & 0.6398$_{\pm0.0210}$ & 0.6975$_{\pm0.0186}$ & 0.6931$_{\pm0.0189}$ & 0.6981$_{\pm0.0165}$ \\
GWO       & 0.5939$_{\pm0.0237}$ & 0.5645$_{\pm0.0209}$ & 0.6150$_{\pm0.0164}$ & 0.6623$_{\pm0.0198}$ & 0.6649$_{\pm0.0166}$ & 0.6626$_{\pm0.0154}$ \\
\rowcolor{black!8}
RLMA      & 0.8697$_{\pm0.0131}$ & 0.7997$_{\pm0.0149}$ & 0.8210$_{\pm0.0078}$ & 0.8304$_{\pm0.0104}$ & 0.8151$_{\pm0.0074}$ & 0.7979$_{\pm0.0097}$ \\
\rowcolor{black!18}
\textbf{RLOSMEA} & \textbf{0.9394}$_{\pm\textbf{0.0052}}$ & \textbf{0.9149}$_{\pm\textbf{0.0122}}$ & \textbf{0.9321}$_{\pm\textbf{0.0050}}$ & \textbf{0.9252}$_{\pm\textbf{0.0114}}$ & \textbf{0.9040}$_{\pm\textbf{0.0250}}$ & \textbf{0.9018}$_{\pm\textbf{0.0264}}$ \\
\bottomrule
\end{tabular}
\end{table*}

\begin{table*}[!t]
\centering
\footnotesize
\caption{Runtime comparison of six algorithms across six scenarios under 10,000 function evaluations. Each value reports the mean $\pm$ standard deviation in seconds over 30 independent runs.}
\label{tab05}
\renewcommand{\arraystretch}{2}
\begin{tabular}{@{}l c c c c c c@{}}
\toprule
Algorithm & scenario\_01 & scenario\_02 & scenario\_03 & scenario\_04 & scenario\_05 & scenario\_06 \\
\midrule
MemeticEA & 25.88$_{\pm1.12}$ & 35.03$_{\pm0.80}$ & 51.75$_{\pm0.79}$ & 69.50$_{\pm0.44}$ & 83.48$_{\pm1.21}$ & 97.51$_{\pm0.94}$ \\
ALNS      & 24.63$_{\pm1.15}$ & 36.21$_{\pm10.34}$ & 48.20$_{\pm1.37}$ & 66.88$_{\pm1.72}$ & 81.25$_{\pm1.97}$ & 96.81$_{\pm2.63}$ \\
AESSPSO   & 35.19$_{\pm1.47}$ & 46.12$_{\pm9.78}$ & 60.23$_{\pm1.29}$ & 79.14$_{\pm1.81}$ & 90.96$_{\pm2.04}$ & 103.48$_{\pm1.30}$ \\
GWO       & 26.55$_{\pm10.25}$ & 31.67$_{\pm1.11}$ & 47.75$_{\pm1.21}$ & 66.34$_{\pm1.80}$ & 78.27$_{\pm1.72}$ & 92.77$_{\pm1.74}$ \\
RLMA      & 27.56$_{\pm0.58}$ & 38.67$_{\pm1.23}$ & 55.40$_{\pm1.65}$ & 74.83$_{\pm1.39}$ & 85.97$_{\pm1.55}$ & 98.45$_{\pm2.05}$ \\
\rowcolor{black!18}
\textbf{RLOSMEA} & \textbf{35.23}$_{\pm\textbf{0.71}}$ & \textbf{51.76}$_{\pm\textbf{0.93}}$ & \textbf{69.19}$_{\pm\textbf{1.36}}$ & \textbf{87.29}$_{\pm\textbf{2.07}}$ & \textbf{104.93}$_{\pm\textbf{2.59}}$ & \textbf{121.03}$_{\pm\textbf{1.99}}$ \\
\bottomrule
\end{tabular}
\end{table*}

Fig.~\ref{fig04} presents the evolution of the three utility components in representative scenarios. RLOSMEA achieves higher task-gain utility while maintaining competitive load-balance performance, indicating that its improvement is not obtained by simply concentrating tasks on a few favorable satellites. The energy-saving utility is not always the highest, which is consistent with the gain-oriented weight setting used in the main comparison. Therefore, the advantage of RLOSMEA should be interpreted as a better global tradeoff among task gain, energy consumption, and load distribution rather than simultaneous dominance in every individual component.

\subsubsection{Statistical Comparison}

Table~\ref{tab04} reports the mean and standard deviation of the weighted utility over 30 independent runs. RLOSMEA obtains the highest mean weighted utility in all tested scenarios. Compared with the strongest RL-assisted baseline, RLMA, RLOSMEA improves the mean weighted utility by 8.01\%, 14.41\%, 13.53\%, 11.42\%, 10.91\%, and 13.02\% from scenario\_01 to scenario\_06, respectively. These results show that the proposed method maintains a consistent advantage under different heterogeneous scheduling scales.

\subsubsection{Runtime Comparison}

Table~\ref{tab05} reports the wall-clock runtime under the same 10,000-function-evaluation budget. The runtime of all algorithms increases with scenario scale. RLOSMEA requires more runtime than most baselines because it performs online operator-mode control and schedule-specific modification before decoder evaluation. Nevertheless, its runtime remains within the same order of magnitude as the compared methods. Together with the weighted-utility results, the runtime comparison indicates that RLOSMEA provides a favorable tradeoff between solution quality and computational cost.

\subsection{Preference-Adjustability Analysis}

\begin{figure}[!h]
	\centering
	\captionsetup[subfigure]{font=scriptsize}
	\subfloat[scenario\_01]{
		\includegraphics[width=0.96\columnwidth]{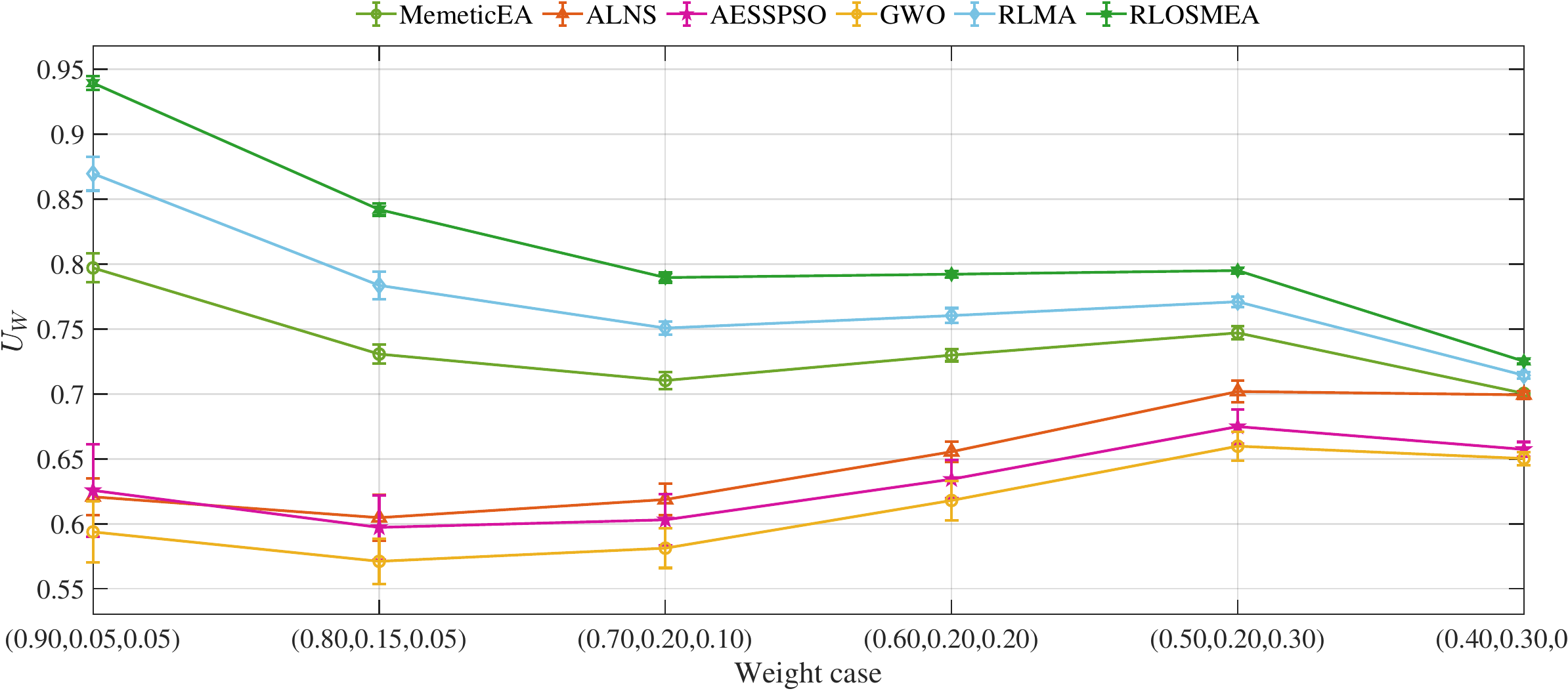}
	}

	\subfloat[scenario\_03]{
		\includegraphics[width=0.96\columnwidth]{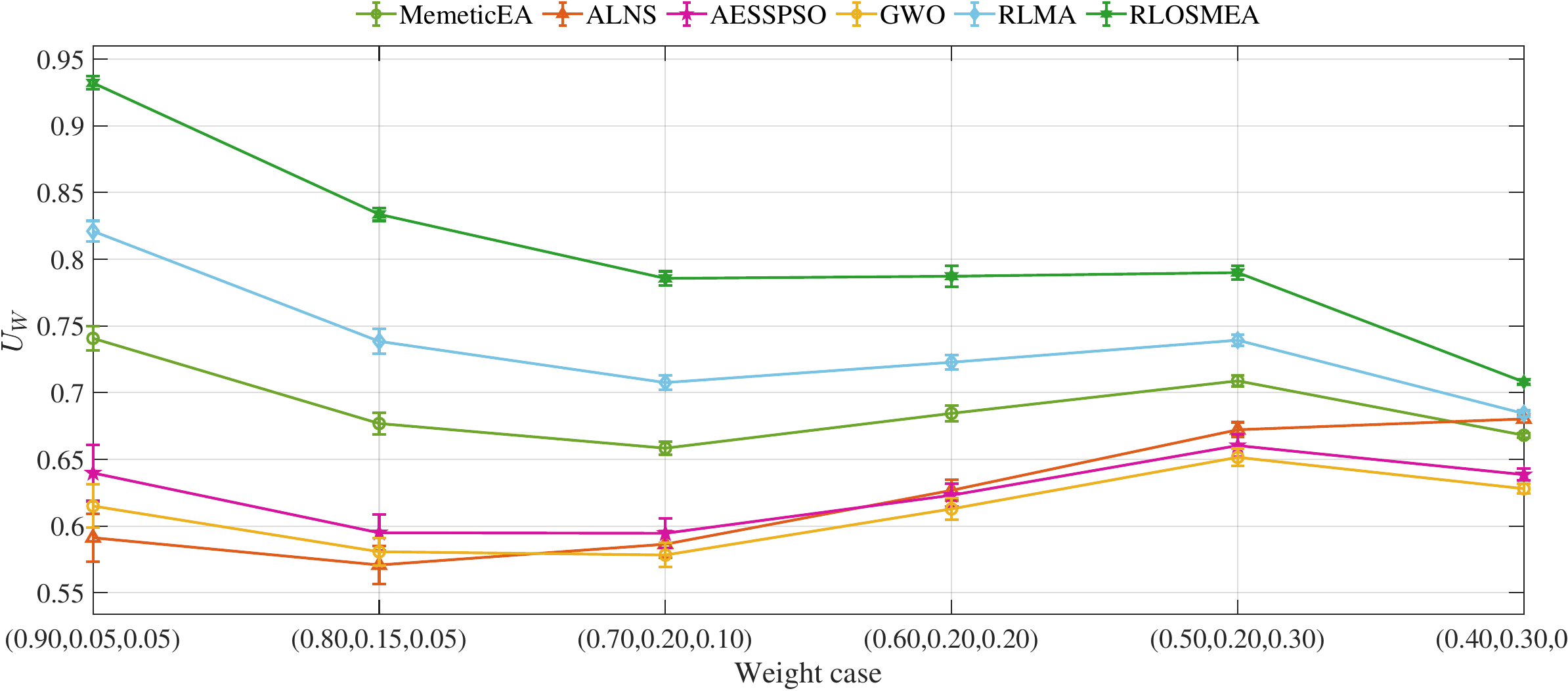}
	}

	\subfloat[scenario\_05]{
		\includegraphics[width=0.96\columnwidth]{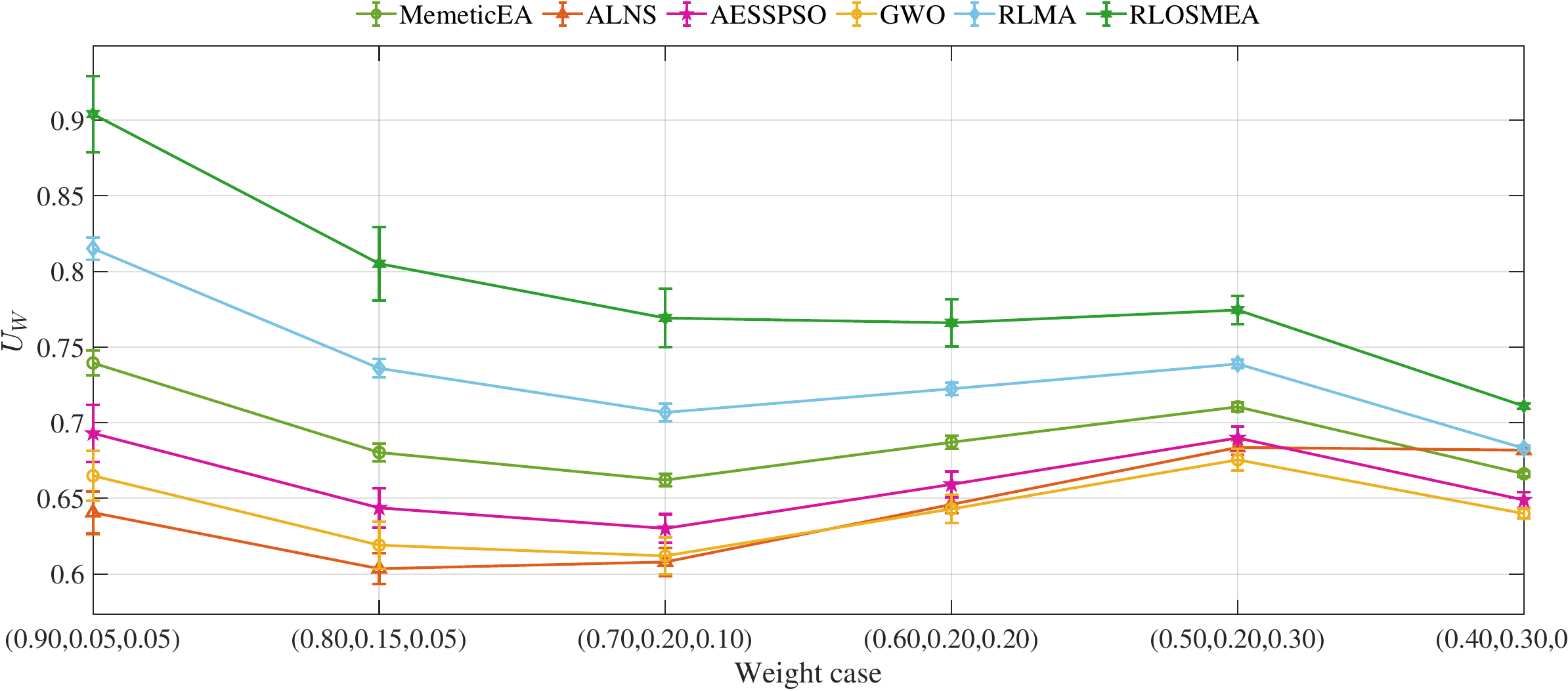}
	}
	\caption{Mean weighted utility under different weight settings in three representative heterogeneous scenarios. The markers and error bars show the mean $\pm$ standard deviation over 30 independent runs.}
	\label{fig05}
\end{figure}

The preference adjustability of the weighted objective is evaluated on representative scenarios with different scheduling scales. The tested weight vectors shift the scheduling preference from gain-dominant operation to more balanced tradeoffs among task gain, energy saving, and load balance. Fig.~\ref{fig05} shows the mean weighted utility under different preference settings. RLOSMEA achieves the best performance across the tested weights and scenarios, indicating that its advantage is not tied to a single manually selected weight vector. The results also show that the proposed method remains stable when the objective emphasis changes, supporting the preference-adjustable property of the scheduling model.

\subsection{Parameter Robustness Analysis}

\subsubsection{Single-Parameter Sensitivity Across scenarios}

\begin{figure}[!h]
	\centering
	\includegraphics[width=0.99\columnwidth]{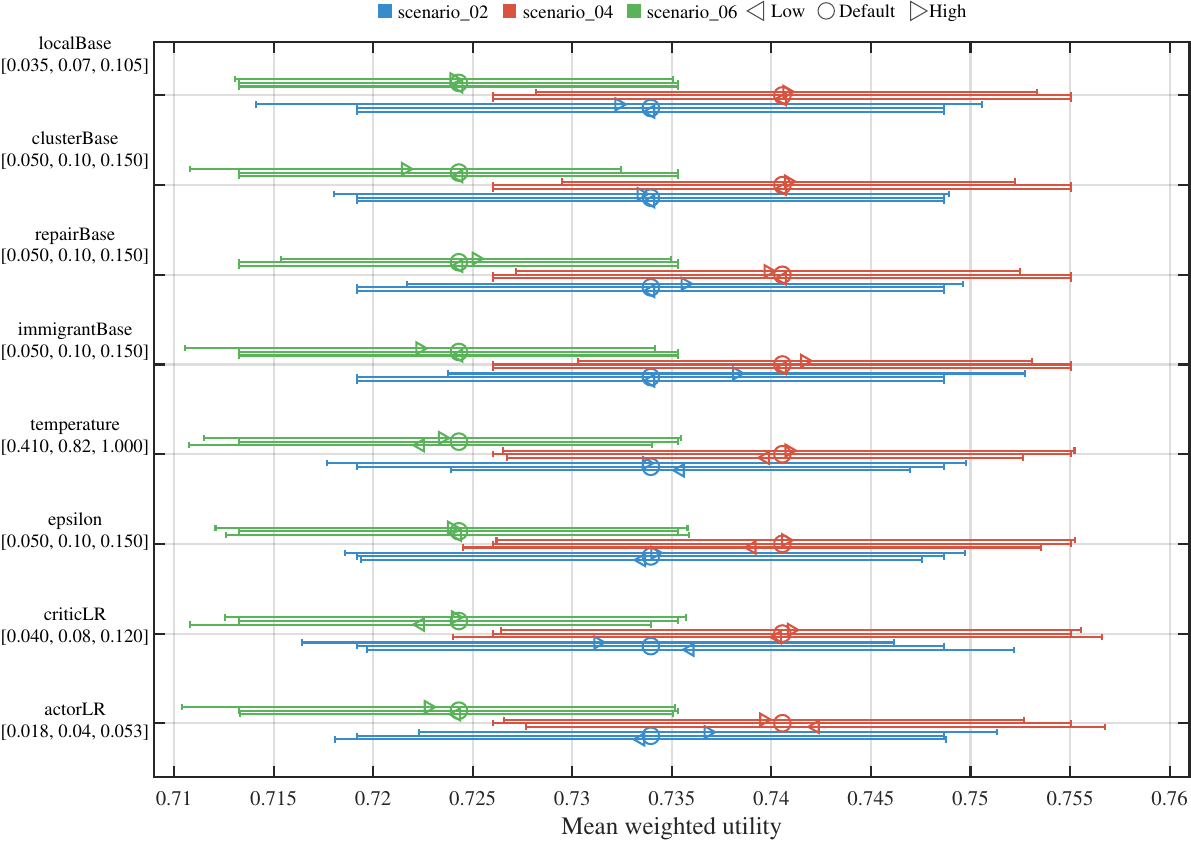}
	\caption{Parameter sensitivity of RLOSMEA across three representative scenarios. For each parameter, three levels (Low, Default, High) are evaluated. The markers and error bars denote the mean $\pm$ standard deviation of the weighted utility over repeated runs.}
	\label{fig06}
\end{figure}

To examine whether the parameter behavior is scenario-dependent, single-parameter sensitivity analysis is conducted on scenario\_02, scenario\_04, and scenario\_06. These scenarios represent different scheduling scales. For each tested parameter, Low, Default, and High levels are evaluated while the other parameters are fixed at their default values. Fig.~\ref{fig06} reports the parameter-sensitivity results. Although the absolute weighted-utility values vary with scenario scale, the relative trends are generally stable across the three scenarios. No abrupt performance degradation is observed within the tested ranges, suggesting that the default parameter setting lies in a stable region rather than near an isolated optimum.

\begin{figure*}[!t]
	\centering
	\includegraphics[width=0.98\textwidth]{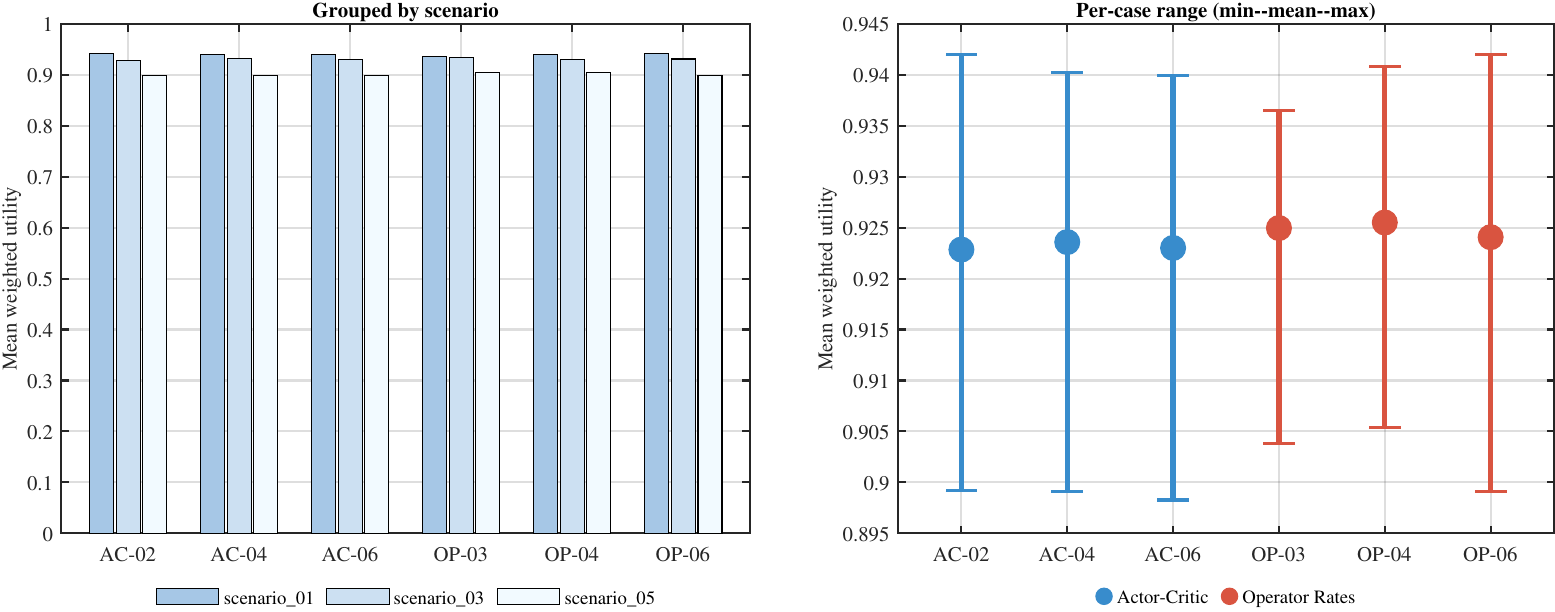}
	\caption{Validation performance of selected coupled parameter cases. AC denotes selected actor--critic parameter cases, and OP denotes selected operator-rate parameter cases. The bars show the mean weighted utility, and the error bars indicate the standard deviation over repeated runs.}
	\label{fig07}
\end{figure*}

\subsubsection{Coupled Parameter Tuning}

To examine coupled parameter effects, a reduced-cost screen-then-validate strategy is adopted. The actor--critic group includes actor learning rate, critic learning rate, exploration rate, and softmax temperature, while the operator-rate group includes immigrant rate, repair rate, conflict-cluster adjustment rate, and local improvement rate. For each group, an L$_9$ orthogonal design is used with three relative levels: 0.5, 1.0, and 1.5 times the default value. The selected candidate cases are then validated on representative scenarios. Fig.~\ref{fig07} shows that the selected actor--critic and operator-rate cases obtain close mean weighted utilities. This indicates that RLOSMEA remains stable under moderate coupled parameter variations and is not dependent on a narrowly tuned configuration.

\subsection{Ablation Study}

To evaluate the contribution of the main components, the full RLOSMEA is compared with four variants: w/o RL Selection, w/o Operator Modes, w/o Phase Descriptor, and w/o Recovery Mode. The w/o RL Selection variant disables RL-based operator selection; w/o Operator Modes removes the six AEOS-oriented operator modes; w/o Phase Descriptor replaces adaptive phase-aware control with a fixed-transition strategy; and w/o Recovery Mode removes the safeguard for severe stagnation or diversity collapse.

\begin{figure*}[!t]
	\centering
	\includegraphics[width=0.9\textwidth]{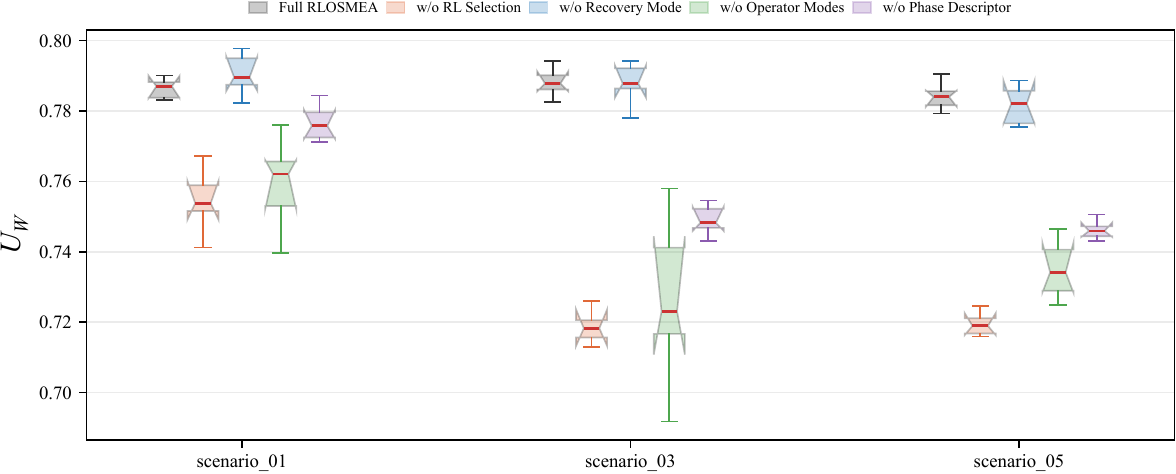}
	\caption{Ablation results of RLOSMEA on three representative heterogeneous scheduling scenarios. The full RLOSMEA is compared with four ablation variants: w/o RL Selection, w/o Recovery Mode, w/o Operator Modes, and w/o Phase Descriptor. The boxplots show the distribution of weighted utility over repeated runs.}
	\label{fig08}
\end{figure*}

Fig.~\ref{fig08} shows the ablation results. Removing RL-based operator selection causes the most significant degradation, especially in scenario\_03 and scenario\_05. Removing AEOS-oriented operator modes also leads to clear performance loss, confirming the importance of gain construction, rare-task shaping, feasibility repair, conflict resolution, elite intensification, and diversity exploration. The w/o Phase Descriptor variant performs worse than the full version, showing that phase-aware control is more effective than a fixed transition strategy. The w/o Recovery Mode variant remains close to the full version in the tested scenarios, which is consistent with its role as a safeguard activated mainly under severe stagnation or diversity collapse.

\subsection{Learning Behavior of Operator Selection}

\begin{figure*}[!t]
	\centering
	\captionsetup[subfigure]{font=scriptsize}
	\subfloat[Optimization progress]{
		\includegraphics[width=0.48\textwidth]{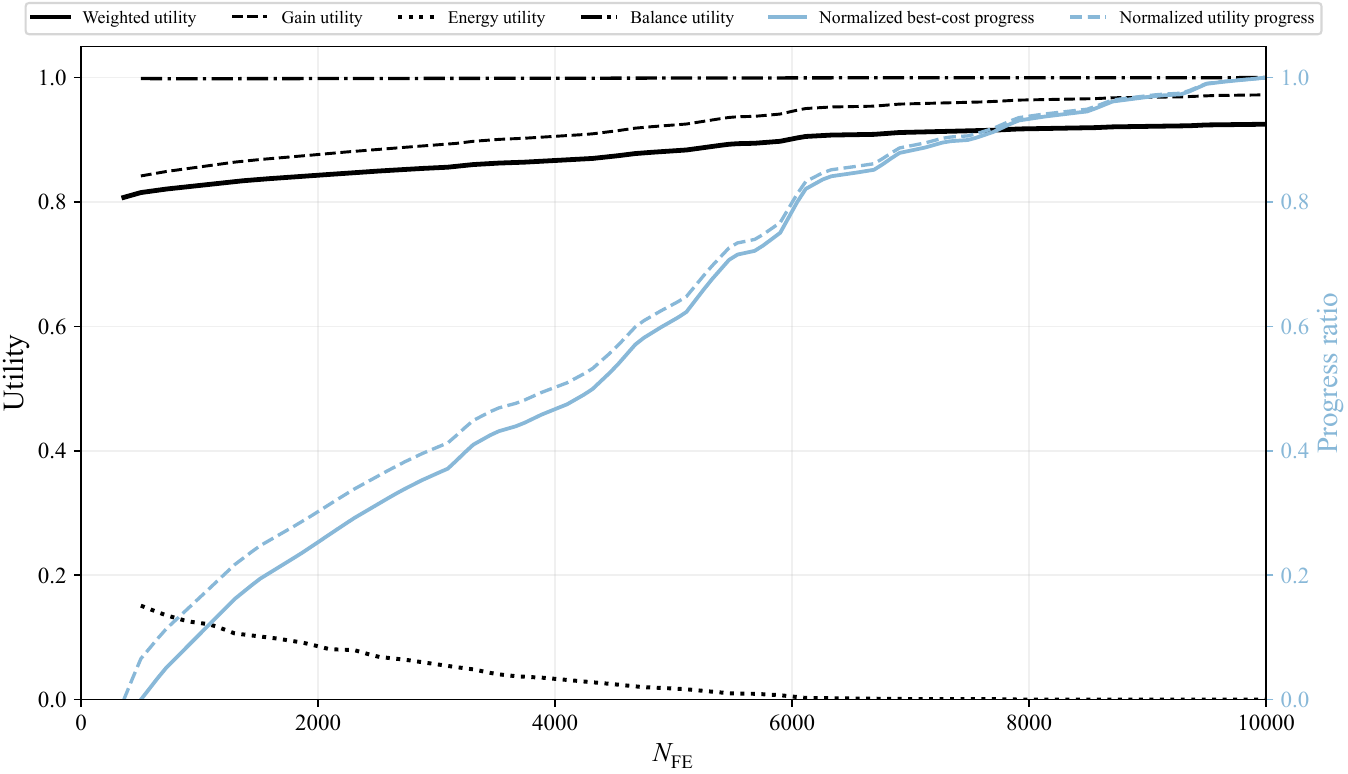}
	}
	\hfill
	\subfloat[Policy action mixture]{
		\includegraphics[width=0.48\textwidth]{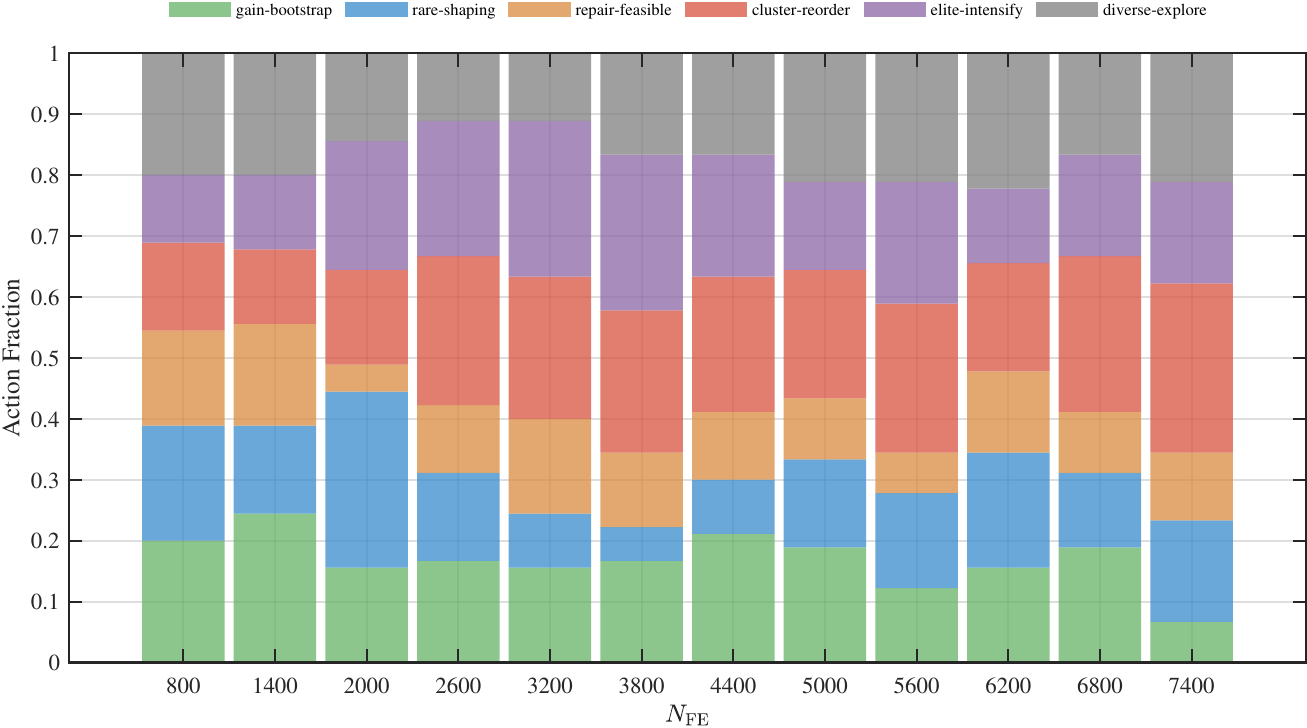}
	}
	\caption{RL diagnostics of RLOSMEA in representative scenario\_04.}
	\label{fig09}
\end{figure*}

Fig.~\ref{fig09} illustrates the learning behavior of the online operator-selection mechanism. Fig.~\ref{fig09}(a) shows the search response under the learned operator policy, and Fig.~\ref{fig09}(b) shows the corresponding policy action mixture. The weighted utility increases steadily during the search, while the gain utility improves and load balance remains at a high level. This indicates that the learned policy improves the scalar objective mainly by capturing more valuable task assignments without substantially degrading workload distribution. The action mixture shows that no single operator mode dominates the entire process. Several modes remain active at different stages, suggesting that the actor--critic controller adaptively coordinates exploration, feasibility recovery, and exploitation rather than degenerating into a fixed operator pattern.

\subsection{Summary}

The experimental results demonstrate the effectiveness and robustness of RLOSMEA for preference-adjustable heterogeneous AEOS scheduling. Across different scenarios, RLOSMEA achieves the highest weighted utility among the compared algorithms and maintains stable convergence. The runtime comparison shows that online operator control introduces additional computational cost, but the runtime remains within the same order of magnitude as the baselines. Preference-adjustability and parameter-robustness analyses show that the method remains stable under different weight settings and parameter configurations. The ablation and learning-behavior results further verify the importance of RL-assisted operator-mode selection, AEOS-oriented operator modes, and phase-aware control.

\section{Conclusion}
\label{sec:conclusion}

This study shows that reinforcement learning can contribute to heterogeneous AEOS scheduling more effectively when it is used to guide evolutionary search behavior rather than to directly construct schedules.
The proposed EPO framework and RLOSMEA achieve this goal by coupling decoder-based schedule evaluation with online actor--critic operator-mode control, enabling the optimizer to preserve satellite-dependent feasibility constraints while adaptively coordinating exploration, feasibility recovery, refinement, and diversity maintenance.
The experimental results confirm that the proposed method attains the intended objectives: it improves weighted utility and convergence stability over the compared algorithms across heterogeneous scenarios, remains effective under different preference settings, and maintains robustness under parameter variation and ablation tests.
These findings indicate that the proposed framework provides additional scientific value by offering a modular interface between learning-based search control and decoder-based combinatorial scheduling, which can be extended to other resource-constrained mission-planning problems.

The current work is still limited to centralized offline scheduling with preprocessed visibility windows, and does not explicitly model dynamic disturbances, cloud-induced visibility uncertainty, inter-satellite communication, cooperative imaging, or data-relay constraints.
Future research will therefore focus on uncertainty-aware and rolling-horizon scheduling, emergency task insertion, online preference adaptation, and communication- or relay-aware cooperative constellation planning, so that the proposed framework can be further applied to operational large-scale Earth observation missions.

\end{document}